\documentclass[journal]{IEEEtran}
\ifCLASSINFOpdf
  \usepackage[pdftex]{graphicx}
\else
\fi
\usepackage{amsmath}
\usepackage{empheq}

\usepackage{booktabs}
\usepackage{multirow}
\usepackage{xcolor}
\usepackage[center]{caption}
\usepackage{longtable}
\usepackage{lscape}
\usepackage{caption}
\usepackage{subcaption}
\usepackage{array}
\usepackage{arydshln}
\usepackage[flushleft]{threeparttable}
\usepackage{tablefootnote}
\usepackage{float}
\usepackage[hyphens]{url}
\usepackage{hyperref}
\usepackage[hyphenbreaks]{breakurl}

\newcolumntype{L}{>{\centering\arraybackslash}m{3cm}}
\newcolumntype{O}{>{\arraybackslash}m{4cm}}

\begin{document}
\title{An embarrassingly simple comparison of machine learning algorithms for indoor scene classification}

\author{Bhanuka Manesha Samarasekara Vitharana Gamage~% <-this % stops a space
\thanks{Bhanuka M.S.V Gamage is with Monash University Malaysia, Jalan Lagoon Selatan, 47500 Bandar Sunway,
Selangor Darul Ehsan, Malaysia.}}

% \markboth{Journal of \LaTeX\ Class Files,~Vol.~14, No.~8, August~2015}%
% {Shell \MakeLowercase{\textit{et al.}}: Bare Demo of IEEEtran.cls for IEEE Journals}

\maketitle

\begin{abstract}

  With the emergence of autonomous indoor robots, the computer vision task of indoor scene recognition has gained the spotlight.
  Indoor scene recognition is a challenging problem in computer vision that relies on local and global features in a scene.
  This study aims to compare the performance of five machine learning algorithms on the task of indoor scene classification to identify the pros and cons of each classifier.
  It also provides a comparison of low latency feature extractors versus enormous feature extractors to understand the performance effects.
  Finally, a simple MnasNet based indoor classification system is proposed, which can achieve 72\% accuracy at 23 ms latency.

\end{abstract}


% \fi
\IEEEpeerreviewmaketitle

\section{Introduction}

\IEEEPARstart{I}{ndoor} scene recognition is a supervised machine learning problem that has its origins deeply rooted in image classification.
The complexity of this task is directly related to the objects and features in the scene.
It is a classification task, that has many uses such as obstacle avoidance for the visually impaired \cite{rodriguez2012obstacle}, autonomous flight drones \cite{tomic2012toward} and autonomous indoor robots such as service vacuums \cite{liu2009scene, breuer2012johnny}.
These use cases need the ability to localize in indoor locations in real-time with minimal latency.

With the advancements in deep learning and transfer learning, indoor scene recognition algorithms increased in performance.
This is mainly due to the ability to transfer the knowledge learnt in one domain where there exists a high volume of data such as ImageNet with 14 million images and apply the knowledge to domain similar problems.
The process of performing feature extraction is crucial to the performance of the indoor scene classifiers.

This paper aims to perform a head to head comparison between the machine learning classifiers when combined with powerful pre-trained feature extractors.
Two main feature extractors are compared to identify the performance difference between low latency MnasNet feature extractor and enormous ResNext feature extractor.
Both feature extractors are pre-trained on the ImageNet dataset without any fine-tuning.

The main contributions of this work include the following:
\begin{itemize}
 \item Comparison of two state-of-the-art feature extractors with low latency vs enormous size on the MIT-67 dataset.
 \item Comparison of machine learning classifiers on the MIT-67 dataset with pre-trained feature extractors.
 \item A novel yet simple architecture for indoor scene classification using MnasNet with very minimal latency and decent performance.
 \item Benchmark and code for future experiments with the MIT-67 dataset.
\end{itemize}

The rest of this paper is organized as follows: 
Section \ref{background} dives into the related studies and formalizes the aim of the study.
In Section \ref{methodology}, the feature extractors and machine learning classifiers used in the study are introduced.
The high-level architecture used in the study, along with the training process, is explained in Section \ref{implementation}.
Finally, the results obtained by the experiments are compared extensively and summarized in Section \ref{results} and \ref{conclusion}.

\section{Background} \label{background}

\subsection{Indoor Scene Classification}

Indoor Scene classification is considered a challenging task in machine learning. 
This is mainly due to the combination of spatial and objects patterns to describe a scene \cite{quattoni2009recognizing}.
For example, a corridor can easily be identified based on the lines and walls, whereas a book store is identified using books on shelves.
Indoor scenes are comparatively harder to classify compared to outdoor scenes due to the complexity of the objects in indoor scenes \cite{tang2017g}.
The earlier methods that used objects for classifying scenes struggled when classifying indoor scenes compared to outside scenes \cite{oliva2001modeling}.

Feature extraction is the most critical step in scene recognition \cite{liu2019novel}. 
Having well-defined features containing global and local features can improve the performance of the classifiers.
Many proposed methods exist that uses different features to perform the classification.
The methods can be classified into two groups, traditional hand-crafted features and Neural Network based features.

\subsubsection{Traditional hand-crafted methods}
A. Quattoni and A. Torralb \cite{quattoni2009recognizing} established the baseline by their proposed model that uses ROI with GIST features to obtain global and local information.
This model was tested on the MIT-67 indoor scene dataset and was able to obtain 25\% accuracy.
The Deformable part-based model (DPM) with latent SVM training, captured salient objects and visual elements combined with global image features was able to achieve 30.40\% accuracy \cite{pandey2011scene}.
The DPM model inspired both hand-crafted and neural network based models.
\cite{centrist} proposed the CENsus TRansform hISTogram (CENTRIST), a visual descriptor that is easier to implement and is faster to evaluate while also performing better than the GIST and SIFT methods.
They were able to achieve 36.8\% accuracy on the MIT-67 dataset.
The GBPWHGO model proposed by \cite{zhou2013scene} uses visual descriptors composed of a GBP (Gradient Binary Pattern) and a WHGO (Weighted Histogram of Gradient Orientation).
It helps the model capture structural and textural properties of images effectively, which in turn provided a 42.9\% accuracy.
\cite{doersch2013mid} proposed a model that extends from the mean shift algorithm to obtain an accuracy of 66.87\%.
The ISPR \cite{lin2014learning} method learns important spatial pooling regions with an appearance which allows them to achieve 50.10\% on a single feature and 68.50\% on multi-feature.
This model was state-of-the-art (SOTA) for hand-crafted features with the performance on the MIT-67 dataset.

\subsubsection{Neural Network based methods}
Multiple methods have been proposed that utilizes neural networks for feature extraction \cite{bai2018categorizing, liu2019novel, tang2017g, cheng2018scene}.
The Restricted Boltzman Machines based CCRBM model was proposed by \cite{gao2016novel}, which reduces sources of instabilities by using centred factors into the learning strategy.
This method was able to achieve an accuracy of 42.1\% on the MIT-67 dataset.
\cite{qi2016dynamic} proposed a method that uses a well trained ConvNet using transfer learning to extract mid-level features from the image to then perform the classification.
The extracted features were then used to train a neural network and a linear SVM.
Experiments were not conducted on the MIT-67 dataset, but they were able to achieve good results on other scene classification datasets.
\cite{tang2017g} used the GoogLeNet model as a feature extractor and was able to achieve 64.48\% accuracy on the MIT-67 dataset by using pre-trained weights on ImageNet \cite{deng2009imagenet} and Places205 \cite{zhou2014learning}. 
Due to the domain similarity of the datasets, the features can transfer well on to the indoor scene classification task. 
With fine-tuning, they were able to achieve an accuracy of 79.63\%, which made this model SOTA at the time.
Nevertheless, with the advancements in deep learning architectures, the ResNet model \cite{he2016deep} has proven to be better than GoogLeNet for image classification task on ImageNet dataset.
\cite{liu2019novel} proposed a novel ResNet based transfer learning model utilizing multi-layer feature fusion and data augmentation which was able to achieve an accuracy of 94.05\% on the MIT-67 dataset.
This was, however, achieved by increasing the number of images using data augmentation by 475\%.
The accuracy without data augmentation was 74.53\%, which is worse than \cite{tang2017g}.
\cite{bai2018categorizing} proposed an architecture which combines CNN, SVM and random forest sequentially along with spectral clustering to obtain an accuracy of 80.75\%.
VSAD model \cite{wang2017weakly} used PatchNet, a patch-level end-to-end model for feature extraction which then aggregates local features and patches with global representation to achieve an accuracy of 86.2\%.
Another method was proposed by \cite{cheng2018scene}, which introduces a new semantic descriptor with objectness method for scene recognition to exploit the correlation of object configurations across scenes.
It uses ImageNet based CNN to extract the local representation and another CNN for global representation.
This method can achieve a final accuracy of 86.76\%, which is the current SOTA approach for the MIT-67 dataset without data augmentation.

\subsection{CNN as feature extractor}

So taking into consideration the two groups, it can be observed that in the pre-deep learning era, most of the features were extracted using hand-crafted methods such as SIFT \cite{lowe2004distinctive}.
However, once the Convolutional Neural Network (CNN) became popular, many models were trained to be used as feature extractors.
Nevertheless, the downside was the training process needed high-end resources and more extended time.
To overcome this issue, many researchers have experimented and analyzed the effect of transfer learning \cite{tang2017g, sharif2014cnn, qi2016dynamic, liu2019novel}.
It was evident that if the model were trained on the same domain as the transferred dataset, with a bit of fine-tuning, the models could converge to better performance.
So off the shelf CNN features trained on ImageNet were used for many recognition tasks as it produced better results.
\cite{sharif2014cnn} perform a series of experiments to observe the effect of transfer learning on many recognition tasks such as Flower recognition and Sculptures retrieval.
For scene recognition, they obtained an accuracy of 58.4\% on the MIT-67 dataset and with data augmentation, they were able to achieve 69\% accuracy.
This led to many researchers using CNN features with transfer learning for scene classification \cite{qi2016dynamic, tang2017g,liu2019novel}.

However, when selecting a feature extractor, it is also essential to consider the latency of the model when performing inference.
With the age of IoT and embedded systems, the importance of having the ability for scene recognition models to make inference on low powered chips is evident.
Currently, most of the research focuses on using server-level hardware to run and test the models.
\cite{zhang2019towards} compared multiple lightweight deep learning models (ResNet-50 \cite{he2016deep}, MnasNet \cite{tan2019mnasnet}, MobileNetV2 \cite{sandler2018mobilenetv2}) on the action recognition task.
They found that the MnasNet model architecture \cite{tan2019mnasnet} had the lowest latency and the best accuracy for mobile devices.
\cite{liu2009scene} proposed a method that uses geometric and colour information to perform scene recognition on robots.
Currently, there exists a gap to explore the effect of MnasNet model architecture on indoor scene recognition.

In summary, Table \ref{table:literature-review-summary} provides the accuracies obtained on the MIT-67 dataset by prior research studies.
Most studies focus on Neural Networks and Support Vector Machines as the classifiers, but is there a reason to consider these classifier techniques over other methods such as K-Nearest Neighbour, Naive-Bayes and Decision Trees.
This can be identified as a gap, to explore and understand the performance of different classifiers when using SOTA feature extractors.
Another gap that exists is the performance comparison between SOTA low latency feature extractors (MnasNet) vs enormous feature extractors with millions of parameters (ResNeXt \cite{xie2017aggregated}) for indoor scene classification.

Therefore this paper aims to answer these two questions,
\begin{enumerate}
 \item What is the performance difference between machine learning classifiers in indoor scene classification?
 \item How well does a SOTA low latency feature extractor compare to an enormous feature extractor in indoor scene classification?
\end{enumerate}

\renewcommand{\arraystretch}{1.2}

\begin{table}
  \centering
  \caption{Accuracy on MIT-67 dataset}
  \label{table:literature-review-summary}
  \resizebox{0.48\textwidth}{!}{
  \begin{tabular}{Llc}
  \hline
  \textbf{ Category }                                & \textbf{Method }                 & \textbf{Accuracy (\%)}  \\ 
  \hline
  \multirow{7}{\linewidth}{\centering Traditional Hand-crafted features} & ROI  \cite{quattoni2009recognizing}                            & 25.0                    \\
                                                     & DPM \cite{pandey2011scene}                               & 30.40                   \\
                                                     & CENTRIST \cite{centrist}                         & 36.8                    \\
                                                     & GBPWHGO \cite{zhou2013scene}                         & 42.9                    \\
                                                     & Mode\-Seeking+IFV  \cite{doersch2013mid}              & 66.87                   \\
                                                     & ISPR \cite{lin2014learning}                             & 50.10                   \\
                                                     & ISPR+IFV \cite{lin2014learning}                         & 68.50                   \\ 
  \hline
  \multirow{9}{\linewidth}{\centering Neural Network based features}     & CCRBM \cite{gao2016novel}                            & 42.1                    \\
                                                     & TCoF \cite{qi2016dynamic}                             & -                       \\
                                                     & CNN-SVM \cite{sharif2014cnn}                          & 58.4                   \\
                                                     & CNNaug-SVM \cite{sharif2014cnn}                          & 69.0                  \\
                                                     & G-MS2F \cite{tang2017g}                          & 79.63                   \\
                                                     & FTOTLM without data augmentation \cite{liu2019novel} & 74.53                   \\
                                                     & FTOTLM with data augmentation \cite{liu2019novel}    & 94.05                   \\
                                                     & VSAD \cite{wang2017weakly}                            & 86.20                   \\
                                                     & SDO \cite{cheng2018scene}                             & 86.76                   \\
  \hline
  \end{tabular}
  }
  \end{table}

  \section{Methodology} \label{methodology}

  This section discusses the dataset, classifiers and methodology used to conduct the experiments.

  \subsection{Dataset}
  
\begin{figure*}
  \centering
  \begin{subfigure}{\linewidth}
    \centering
    \includegraphics[width=\textwidth]{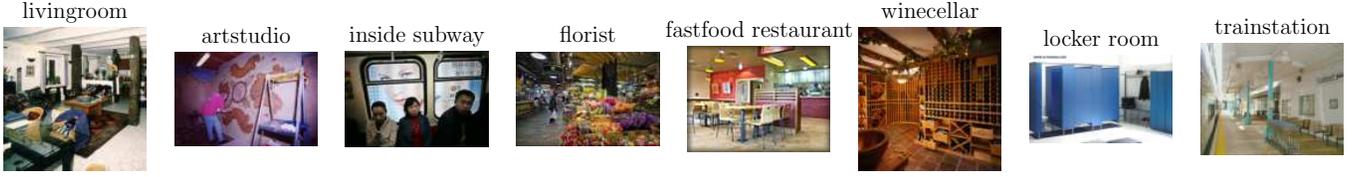}
    \caption{Sample Images from the MIT-67 dataset}
    \label{fig:sampleimages}
  \end{subfigure} 
  \begin{subfigure}{\linewidth}
    \centering
    \includegraphics[width=\textwidth]{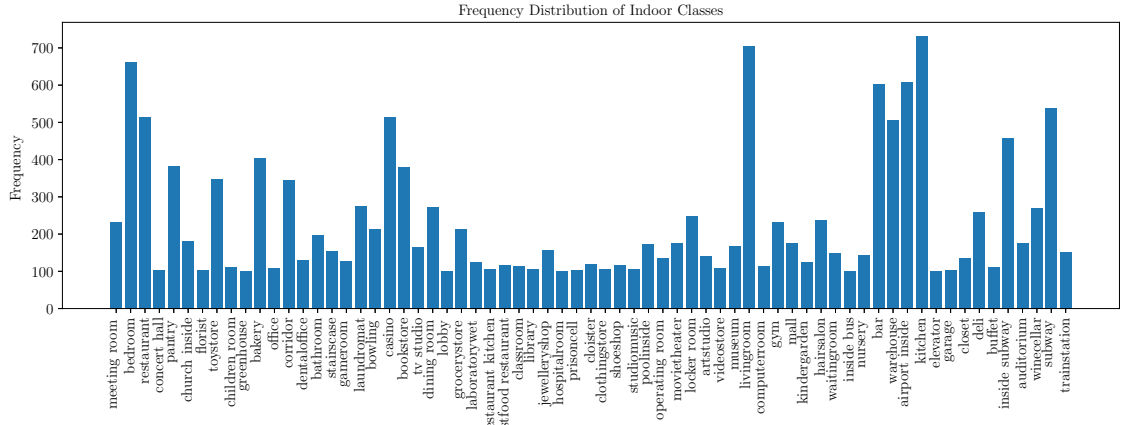}
    \caption{Frequency Distribution of the Classes}
    \label{fig:imagefrequencydist} 
  \end{subfigure}     

  \caption{MIT-67 dataset \cite{quattoni2009recognizing}}
  \label{fig:datasetdetails}

\end{figure*}

  There are many datasets that focus on scene classification such as SUN database \cite{xiao2010sun}, MIT-67 dataset \cite{quattoni2009recognizing}, Places205 dataset \cite{zhou2014learning} and Places365 dataset \cite{zhou2017places}.
  However, for indoor scene classification, the MIT-67 dataset introduced in 2009 is the \emph{de facto} dataset.
  It contains 15620 images in 67 classes with each class containing 101-734 images.
  Figure \ref{fig:sampleimages} shows sample images from the dataset.
  The main challenge of the dataset is that the classification depends on the intricate details of the objects and scenes.
  
  The dataset has a class imbalance problem, which is evident in Figure \ref{fig:imagefrequencydist} when plotting the frequency distribution of each class.
  To overcome this, \cite{quattoni2009recognizing} selected 80 random images per class for training and 20 images per class for testing.
  To maintain the consistency in comparison, this study will use the same train and test split resulting in 5360 training images and 1340 test images.
  In order to maximize the training samples, the test split will be used as the validation split for all the experiments.

\begin{figure*}[t]
  \centering 
  \includegraphics[width=\textwidth]{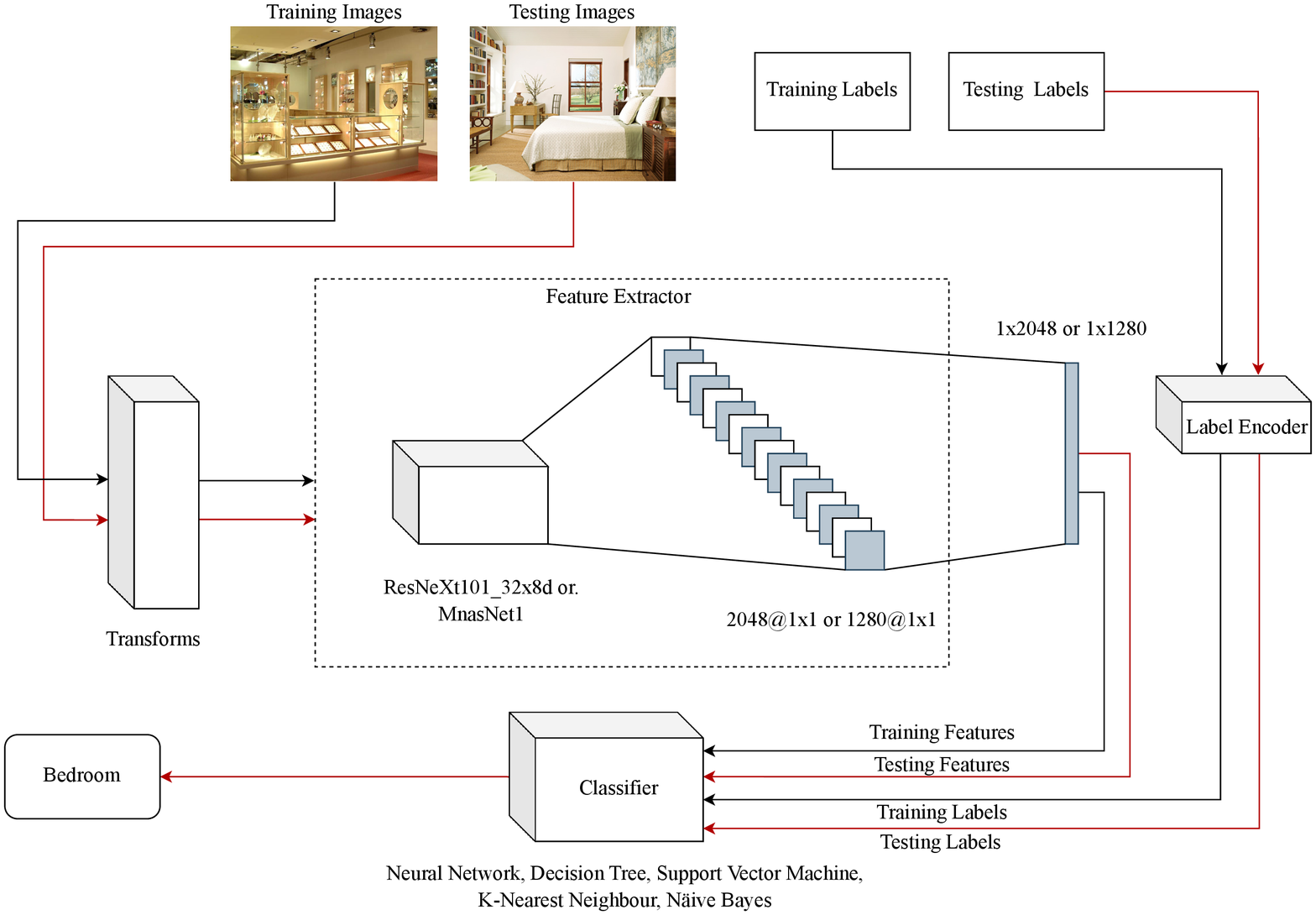}
  \caption{High Level Architecture}
  \label{fig:highlevel}
\end{figure*}

\subsection{Feature Extraction}

Intending to compare the results between the light-weight and high-end SOTA CNN models for feature extraction, the ResNeXt and MnasNet architectures are the choices for the experiments.
In this study, pre-trained weights on ImageNet \cite{deng2009imagenet} dataset is used by both models with no fine-tuning done to the models.
However, \cite{tang2017g, liu2019novel} proved that there are performance gains to be made by fine-tuning on scene classification datasets such as Places205 \cite{zhou2014learning}.
The aim is to understand off the shelf performance on indoor scene classifiers when used as a feature extractor.

\subsubsection{ResNeXt}

The ResNeXt-101 is the current SOTA architecture for ImageNet classification with over 88M parameters.
The ResNeXt-101 (32x48d) version of the model contains 829M parameters and achieves an accuracy of 85.4\% on the ImageNet dataset \cite{mahajan2018exploring}.
For this study, the ResNeXt-101 (32x8d) version pre-trained on ImageNet is used, which contains 88M parameters.
It can achieve 82.2\% Top-1 accuracy on the ImageNet dataset.
In order to obtain the features, this study extracts features before the global \emph{avgpool} layer, which is of shape \emph{1x1x2048}.
This layer is then reshaped to a vector of size \emph{1x2048} to be used as input for the classifiers.

\subsubsection{MnasNet}

The current SOTA model for low latency inference is the MnasNet architecture \cite{tan2019mnasnet}.
It contains 3.9M parameters and achieves 75.2\% Top-1 accuracy on the ImageNet dataset.
Similar to ResNeXt, this study extract the features before the global \emph{avgpool} layer, which is of shape \emph{1x1x1280}.
This is then reshaped to a vector of size \emph{1x1280} to be used as input for the classifiers.

\subsection{Algorithms}

The following five main supervised learning classifiers are tested in order to evaluate their performance.

\subsubsection{K-Nearest Neighbour (KNN)}

A fundamental and straightforward classifier that uses the proximity of the data points to the neighbours to classify the data.
Many distance functions are available; however, the Euclidean distance is used in this study to calculate the distance between the points.
The Euclidean distance between two points $p$ and $q$ is defined as:
\begin{equation}
d(p, q) = \sqrt{(p_{1}-q_{1})^2 + (p_{2}-p_{2})^2 + \cdots + (p_{n}-q_{n})^2}
\end{equation}

\subsubsection{Naive Bayes}

This is a classification technique based on the Bayes Theorem.
It assumes that all the predictors are independent of each other.
Equation \ref{eq:naivebayes} is used to calculate posterior probability.
Even though the technique is simple, it is known to have performance better than complicated classifiers.

\begin{equation}
 p(C_k | x) = \frac{p(C_k)p(x|C_k)}{p(x)}
 \label{eq:naivebayes}
\end{equation}

\subsubsection{Decision Tree}

This is a popular algorithm for classification of both categorical and continuous variables.
Decision trees are beneficial as it provides the ability to interpret the model.
Due to this, decision trees are highly useful in mission-critical application where black-box classifiers are not used.
Different techniques, such as Gini Impurity and Information Gain, are used to divide the data into heterogeneous groups.
The downside to decision trees is that they can be highly biased.

To reduce this, an ensemble of decision trees known as a Random Forest is used.
This study also experiment with random forests to see whether there are any performance gains compared to decision trees.
The downside of a random forest is that the interpretability of the model is lost.

\subsubsection{Support Vector Machine (SVM)}

This is one of the robust classifiers that is commonly used in scene classification.
In most cases, SVM models can match or outperform neural networks.
In the past, SVM was limited to linear classification.
Then, the kernel trick was proposed by \cite{boser1992training} in 1992 that allowed the SVM to create nonlinear classifiers.
In this study multiple kernels such as Linear kernel, Polynomial kernel and Radial Basis Function (RBF) kernel are tested.

\subsubsection{Neural Network (NN)}

\begin{figure}[H]
  \centering
  \begin{subfigure}{0.38\linewidth}
  \includegraphics[width=\linewidth]{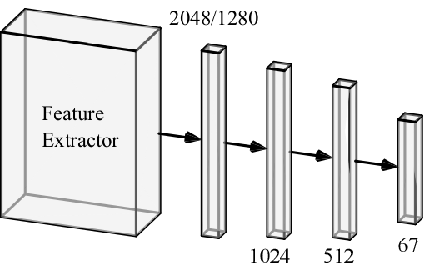}%
  \caption{Shallow Neural Network}%
  \label{fig:shallownn}
  \end{subfigure}
  \quad
  \begin{subfigure}{0.55\linewidth}
  \includegraphics[width=\linewidth]{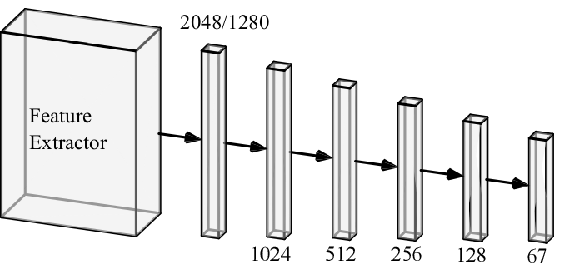}%
  \caption{Deep Neural Network}%
  \label{fig:deepnn}%
  \end{subfigure} 
  \caption{Neural Network architectures used in the experiment}
\end{figure}

This is the final type of classifier, which is widely used in the last few years in indoor scene classification.
The current SOTA architecture also uses neural networks as the classifier.
In order to compare the difference between the shallow and deep neural networks, an additional set of experiments are performed with two different architectures.
Figure \ref{fig:shallownn} and \ref{fig:deepnn} shows the architectures used in the experiments.
The \emph{LogSoftmax} activation function (Equation \ref{eq:logsoftmax}) is used at the final layer, where $x$ vector is a of length $j$ and $i$ is the index of the class.
The \emph{Cross Entropy loss} (Equation \ref{eq:loss}) is used as the loss function, where $y$ is the ground truth and $\hat{y}$ is the prediction and $N$ is the number of items.

\begin{equation}
 LogSoftmax(x_i) = \log{\frac{{\rm e}^{x_i}}{\sum_j{\rm e}^{x}}}
 \label{eq:logsoftmax}
\end{equation}

\begin{equation}
 CrossEntropyLoss = - \frac{1}{N}\left(\sum_i^N{y_i} \log{\hat{y_i}}\right)
 \label{eq:loss}
\end{equation}

 A learning rate decay of $0.96$ is done every ten epochs to reduce the learning rate when achieving convergence.
 This helps the model converge into the global minima quickly, compared to having a higher learning rate which can stop the model from reaching the optimal point.

When training the neural networks, the epoch is set to 500 with early stopping criterion.
The stopping criterion is, if the test accuracy does not increase for 25 epochs, the model will stop training.
For each neural network experiment, the model with the best train accuracy, best test accuracy and the lastest/last model checkpoints are saved.
This allows for the retrieval of the best version of the model for inference.

\begin{table}
  \centering
  \caption{Parameters tested for each classifier}
  \label{tab:my-table}
  
  \end{table}

\begin{table*}[!t]
  \begin{threeparttable}
  \centering
  \caption{Parameters tested for each classifier\tnote{1}}
  \label{tab:parameters}
  \begin{tabular}{l:lOl} 
    \hline
    \multicolumn{1}{l}{\textbf{Classifier}} & \textbf{Parameter}   & \textbf{Description}                                                             & \textbf{Variables}                                                                                                                                                                                                                           \\ 
    \hline
    \multirow{2}{*}{KNN}                    & Algorithm            & Algorithm used to calculate the nearest neighbours                               & \begin{tabular}[c]{@{}l@{}}\begin{tabular}{@{\labelitemi\hspace{\dimexpr\labelsep+0.5\tabcolsep}}l}ball\_tree - Ball Tree\\kd\_tree - K Dimensional Tree\\brute - Brute Force Search\end{tabular}\end{tabular}                               \\ 
    \cdashline{2-4}
                                            & Weights              & Weight function used                                                          & \begin{tabular}[c]{@{}l@{}}\begin{tabular}{@{\labelitemi\hspace{\dimexpr\labelsep+0.5\tabcolsep}}l}uniform - Uniform weights\\distance - Weight points by the inverse of their distance\end{tabular}\end{tabular}                            \\ 
    \hline
    Naive Bayes                             & Var smoothing        & Adds value to the distribution variance to widen the curve                         & \begin{tabular}[c]{@{}l@{}}\begin{tabular}{@{\labelitemi\hspace{\dimexpr\labelsep+0.5\tabcolsep}}l}1e-9\\1e-6\\1e-3\end{tabular}\end{tabular}                                                                                                \\ 
    \hline
    \multirow{2}{*}{SVM}                    & Kernel               & The kernel type used in the algorithm                                            & \begin{tabular}[c]{@{}l@{}}\begin{tabular}{@{\labelitemi\hspace{\dimexpr\labelsep+0.5\tabcolsep}}l}linear - Linear kernel\\poly - Polynomial kernel\\rbf -~Radial Basis Function kernel\\sigmoid - Sigmoid kernel\end{tabular}\end{tabular}  \\ 
    \cdashline{2-4}
                                            & C                    & Regularization parameter                                                         & \begin{tabular}[c]{@{}l@{}}\begin{tabular}{@{\labelitemi\hspace{\dimexpr\labelsep+0.5\tabcolsep}}l}1e-4\\1e-2\\1\\1e2\\1e4\end{tabular}\end{tabular}                                                                                         \\ 
    \hline
    \multirow{2}{*}{Decision Tree}          & Criterion            & The function to measure the quality of a split                                   & \begin{tabular}[c]{@{}l@{}}\begin{tabular}{@{\labelitemi\hspace{\dimexpr\labelsep+0.5\tabcolsep}}l}gini - Gini Impurity\\entropy - Information Gain\end{tabular}\end{tabular}                                                                 \\ 
    \cdashline{2-4}
                                            & Max Depth            & The maximum depth of the tree                                                    & \begin{tabular}[c]{@{}l@{}}\begin{tabular}{@{\labelitemi\hspace{\dimexpr\labelsep+0.5\tabcolsep}}l}10\\50\\100\\None - All leaves are pure\end{tabular}\end{tabular}                                                                                \\ 
    \hline
    \multirow{3}{*}{Random Forest}          & Number of Estimators & Number of trees in the forest                                                    & \begin{tabular}[c]{@{}l@{}}\begin{tabular}{@{\labelitemi\hspace{\dimexpr\labelsep+0.5\tabcolsep}}l}2\\10\\100\\1000\end{tabular}\end{tabular}                                                                                                \\ 
    \cdashline{2-4}
                                            & Criterion            & The function to measure the quality of a split                                   & \begin{tabular}[c]{@{}l@{}}\begin{tabular}{@{\labelitemi\hspace{\dimexpr\labelsep+0.5\tabcolsep}}l}gini - Gini Impurity\\entropy - Information Gain\end{tabular}\end{tabular}                                                                 \\ 
    \cdashline{2-4}
                                            & Max Depth            & The maximum depth of the tree                                                    & \begin{tabular}[c]{@{}l@{}}\begin{tabular}{@{\labelitemi\hspace{\dimexpr\labelsep+0.5\tabcolsep}}l}10\\50\\100\\None - All leaves are pure\end{tabular}\end{tabular}                                                                                \\ 
    \hline
    \multirow{5}{*}{Neural Network}         & Batch Size           & The size of batch used in training and inference                                 & \begin{tabular}[c]{@{}l@{}}\begin{tabular}{@{\labelitemi\hspace{\dimexpr\labelsep+0.5\tabcolsep}}l}64\\32\end{tabular}\end{tabular}                                                                                                          \\ 
    \cdashline{2-4}
                                            & Learning Rate        & The step size at each iteration while moving toward a minimum of the loss function & \begin{tabular}[c]{@{}l@{}}\begin{tabular}{@{\labelitemi\hspace{\dimexpr\labelsep+0.5\tabcolsep}}l}1e-7\\1e-5\\1e-3\\\end{tabular}\end{tabular}                                                                                              \\ 
    \cdashline{2-4}
                                            & Optimizer            & Optimizer used to update the weights to reduce the loss                        & \begin{tabular}[c]{@{}l@{}}\begin{tabular}{@{\labelitemi\hspace{\dimexpr\labelsep+0.5\tabcolsep}}l}adamax -~AdaMax optimizer \cite{kingma2014adam} \\adam - Adam optimizer \cite{kingma2014adam} \\sgd - Stochastic Gradient Descent optimizer \cite{robbins1951stochastic} \end{tabular}\end{tabular}                                             \\ 
    \cdashline{2-4}
                                            & Architecture Type    & Shallow FCN architecture\tnote{2} vs Deep FCN architecture\tnote{3}                                 & \begin{tabular}[c]{@{}l@{}}\begin{tabular}{@{\labelitemi\hspace{\dimexpr\labelsep+0.5\tabcolsep}}l}shallow - 3 layers\\deep - 5 layers\end{tabular}\end{tabular}                                                                             \\ 
    \cdashline{2-4}
                                            & Learning Rate Decay \cite{you2019does}  & Decay the learning rate                                                          & \begin{tabular}[c]{@{}l@{}}\begin{tabular}{@{\labelitemi\hspace{\dimexpr\labelsep+0.5\tabcolsep}}l}wd - Performs learning rate decay\\(no wd) - Do not perform learning rate decay\end{tabular}\end{tabular}                                 \\
    \hline
    \end{tabular}
  \begin{tablenotes}
    \small
    \item[1] All the experiments are repeated for both the feature extractors.
    \item[2] Shallow FCN contains a Neural Network with three linear layers as shown in Figure \ref{fig:shallownn}
    \item[3] Deep FCN contains a Neural Network with five linear layers as shown in Figure \ref{fig:deepnn}] 
  \end{tablenotes}
  \end{threeparttable}
  \end{table*}

  \subsection{Evaluation methods}

  There are multiple evaluation metrics to compare the performance of classifiers.
  The primary metric used on the MIT-67 dataset is the accuracy (Equation \ref{eq:accuracy}) of the model.
  Since the dataset is balanced, the accuracy of the model is a good representation of the performance of the classifier.
  However, the precision, recall and F1 score (Equation \ref{eq:f1score}) is calculated for each class which is then used to calculate the weighted average for all the metrics.
  Precision and Recall are calculated using Equation \ref{eq:precision} and \ref{eq:recall} respectively, using the True Positive (TP), True Negative (TN), False Positive (FP) and False Negative (FN) values. 
  Finally, the Receiver Operating Characteristic (ROC) curve is plotted for each class along with the confusion matrix to visualize the performance easily.
  However, for this study, the accuracy and inference time are the key metrics used to analyze the performance of the classifiers.
  
  \begin{empheq}[left=\text{Metrics} \empheqlbrace]{align}
    & \label{eq:accuracy} \text{Accuracy} = \frac{TP + TN}{TP + TN + FP + FN} \\ 
    & \label{eq:precision} \text{Precision} = \frac{TP}{TP + FP} \\ 
    & \label{eq:recall} \text{Recall} = \frac{TP}{TP + FN} \\ 
    & \label{eq:f1score} \text{F1-Score} = 2 \times \frac{\text{Precision}\times \text{Recall}}{\text{Precision} + \text{Recall}}
    \end{empheq}    

  In order to evaluate the speed, the stages of extracting features, training and inference are timed and logged in seconds.
  When calculating the total time taken $T$ for the train and test datasets, the feature extraction time $FE$ is added to the training time $x_{train}$ and testing time $y_{test}$ and the inference time is divided by the number of images $n$, as shown in Equation \ref{eq:traintime} and \ref{eq:testtime}.
    
\begin{empheq}[]{align}
  & \label{eq:traintime} T_{train} = FE_{train} + x_{train} \\ 
  & \label{eq:testtime}  T_{test} = \frac{FE_{test} + x_{test}}{n_{test}}
  \end{empheq}

\section{Implementation} \label{implementation}

\subsection{Data Pre-processing}

Before training the model, the data is preprocessed.
First, as the image sizes are not constant in the MIT-67 dataset, the images are resized to \emph{224x224} to be passed into the feature extractors.
Next, the images are converted to tensors to be used by the machine learning frameworks.
Finally, the image is normalized to the mean and standard deviation of the ImageNet dataset.
This is done as the pre-trained weights of the models are trained on the ImageNet dataset.
This ensures that the images are normalized to the same distribution as the ImageNet.

All the image labels are converted into numerical values between 0 and 66 using the sklearn LabelEncoder.
Then the numerical representation is converted to One Hot Encoding when performing the training and evaluation process.

\subsection{Training}

All the experiments were done on a PC running \emph{Ubuntu 18.04.1 LTS} with an \emph{Intel® Xeon(R) W-2145 CPU @ 3.70GHz} with 16 logical cores and 64GB RAM.
A single \emph{Quadro P5000} GPU was used to perform the neural network training with a 7200RPM \emph{Seagate} hard drive to store the data.
Parallel training is done on 16 threads on classifiers that support it, such as K-NN and Random Forest.

Figure \ref{fig:highlevel} shows the high-level architecture and flow of the experiment.
The feature extraction process is done separately to extract the ResNeXt and MnasNet features.
Next, the classifiers are trained using these features.
Finally, the test features are used to infer and calculate the performance of the models.

Multiple training experiments are performed with different parameters.
Table \ref{tab:parameters} shows all the combination of parameters tested.
The experiment name is used to display the parameters used by the classifier.
For example, the neural network with the experiment name \emph{resnext101-shallow-32-adamax-1e-05-wd} was trained on the features extracted through the ResNeXt feature extractor, with the shallow NN architecture, with batch-size 32, using AdaMax optimizer with a learning rate of 1e-05 and with learning rate decay.
All the experiments are repeated with the two feature extractors.
In total, 281 tests were conducted with different parameters, classifiers and feature extractors.

\section{Results and Discussions} \label{results}

All the features, results (log files, metric logs), plots (ROC curves, confusion matrix) and trained models can be obtained using \href{https://www.dropbox.com/s/uc99sliqhotsylh/Results%26Features.zip?dl=0}{\emph{this link}}. 
The results and discussions section focuses on the key observations, however the results allow for more comparisons in the future.
Appendix \ref{appx:results} contains the Accuracy, F1 Score, Time Taken for both train and test dataset on all 281 experiments.

\subsection{ResNeXt vs MnasNet}

The highest accuracy obtained by ResNeXt feature extractor based model is the neural network classifier \emph{resnext101-shallow-64-adamax-1e-05-wd}, with an accuracy of 76.87\% and an inference time of 0.0933 seconds per image.
Similarly, for the MnasNet feature extractor based model it is also a neural network classifier \emph{mnasnet1\_0-shallow-64-adamax-0.001-wd}, with an accuracy of 72.82\% with an inference time of 0.0231 seconds per image.
Comparing the parameters, it is evident that everything is the same except for the learning rate.
The best model for MnasNet was achieved by only training the shallow neural network for three epochs compared to 470 epochs for the ResNeXt based classifier.
This indicates that the MnasNet features allowed the shallow network to converge to its best accuracy quickly.

Comparing the inference times between the MnasNet and ResNeXt, it is evident that there is, on average, 300\% speed increase when using the MnasNet feature extractor.
However, when comparing the accuracies, there is only a 15\% performance decrease in MnasNet compared to ResNeXt.
This shows that the lightweight, low latency network can achieve decent performance, which maintaining low latency compared to the 88M parameter ResNeXt.
Figure \ref{fig:allres} shows this gap in the inference time between the MnasNet and ResNeXt based classifiers.

\begin{table}
  \centering
  \caption{Top accuracy for each classifier and feature extractor}
  \begin{tabular}{p{0.2\linewidth}p{0.4\linewidth}p{0.1\linewidth}p{0.1\linewidth}} 
    \toprule
    Classifier             & Experiment Name                                 & Test Acc. & Test Time  \\ 
    \toprule
    KNN                    & \textit{resnext101-brute-distance}              & 0.6575        & 0.0923     \\
                           & \textit{mnasnet1\_0-ball\_tree-distance}        & 0.6037        & 0.0223     \\ 
    \hline
    Naive Bayes            & \textit{resnext101-1e-09}                       & 0.7187        & 0.0926     \\
                           & \textit{mnasnet1\_0-0.001}                      & 0.6515        & 0.0218     \\ 
    \hline
    Decision Tree          & \textit{resnext101-gini-None}                   & 0.3786        & 0.0922     \\
                           & \textit{mnasnet1\_0-gini-100}                   & 0.2448        & 0.0216     \\ 
    \hline
    Random Forest          & \textit{resnext101-1000-entropy-100}            & 0.7194        & 0.0926     \\
                           & \textit{mnasnet1\_0-1000-gini-100}              & 0.6903        & 0.0221     \\ 
    \hline
    Support Vector Machine & \textit{resnext101-poly-100.0}                  & 0.7672        & 0.1015     \\
                           & \textit{mnasnet1\_0-linear-0.01}                & 0.7276        & 0.0280     \\ 
    \hline
    Neural Network         & \textit{resnext101-shallow-64-adamax-1e-05-wd}  & 0.7687        & 0.0933     \\
                           & \textit{mnasnet1\_0-shallow-64-adamax-0.001-wd} & 0.7299        & 0.0230     \\
    \bottomrule
    \end{tabular}
    \label{tab:topaccuracy}
    \end{table}

    \subsection{All experiments}

    Analyzing all the classifiers, the performance can be ranked based on the Test Accuracy and Inference Time, as shown in Appendix \ref{appx:results}.
    Table \ref{tab:topaccuracy} extracts the top results from each feature extractor and classifier.
    This indicates that some classifiers tend to overfit on the training data, mainly the Decision Tree, thus reducing the test performance.
    
    The classifier performance can be separated into two groups; \emph{Good Performance:} Naive Bayes, Random Forest, Support Vector Machines, Neural Network; \emph{Poor Performance:} Decision Tree, KNN.
    Based on the training time, the faster and simpler model is the Naive Bayes, with only a 7\% drop in performance compared to the neural network classifier.
    This was explored by \cite{saritas2019performance} where there was only a 4\% drop in performance.
    This also means that the training data of 5360 samples may not be enough for a neural network to learn the data distribution thoroughly.
    \cite{liu2019novel} proved this by increasing the accuracy from 74.63\% to 94.05\% by augmenting the sample size to 74,149.
    
\begin{figure}[t]
  \centering
  {\includegraphics[width=\linewidth]{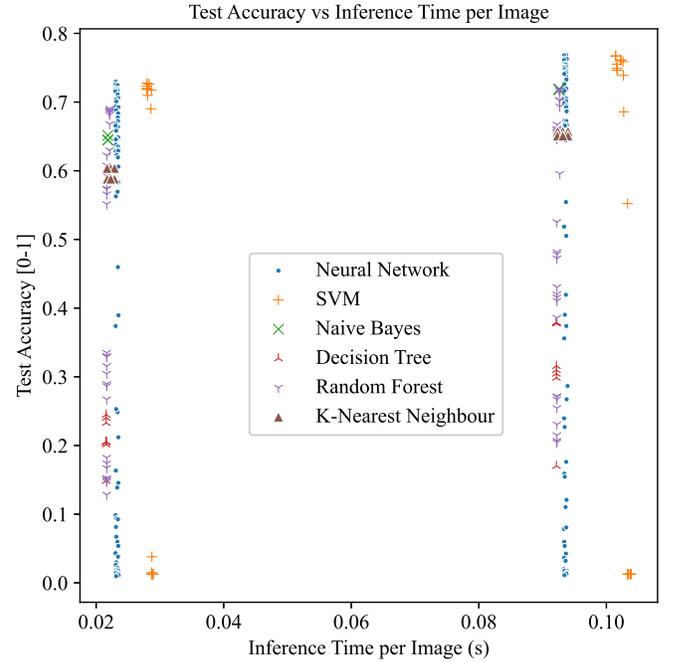}}
  \caption{Comparison of Test Accuracy vs Inference time of all classifiers}
  \label{fig:allres} 
\end{figure}

    Inspecting Figure \ref{fig:allres} it is evident that SVM models take more time for inference compared to all the other classifiers.
    There is an 8\% increase in inference time compared to Naive Bayes, which is also running on the CPU with no parallel execution.
    Therefore, SVM may not be suitable for real-time inference compared to the other classifiers.
    
    Considering all the experiments, the fastest experiment is \emph{mnasnet1\_0-shallow-64-adamax-0.001-wd}, which has an accuracy of 72.9\% with a 23ms latency.

\begin{figure*}[ht]
  \centering
  {\includegraphics[width=\linewidth]{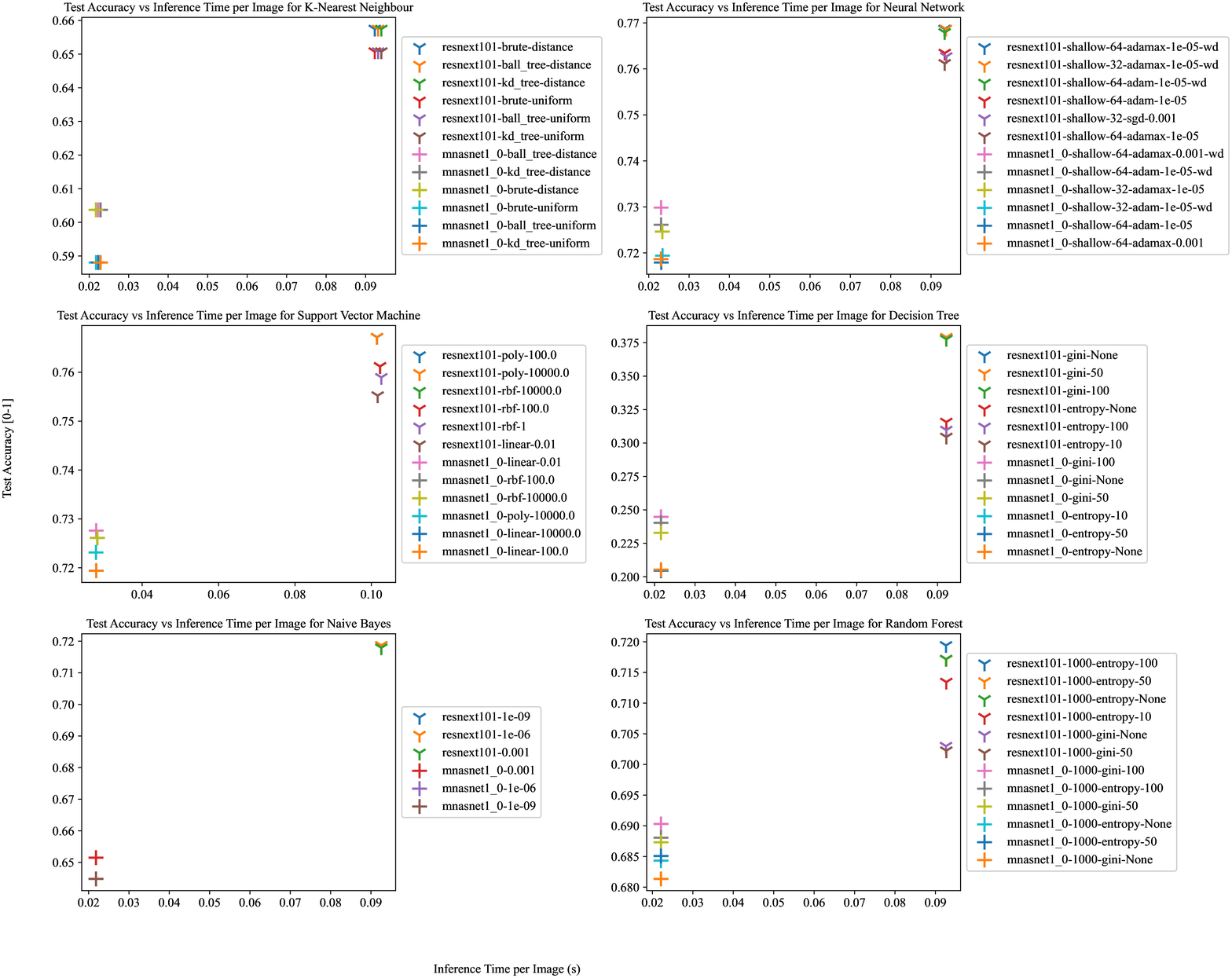}}
  \caption{Top six experiments per classifier and feature extractor}
  \label{fig:classifiertestresults}
\end{figure*}

\subsection{Classifiers}

Next, a comparison of performance based on different parameters used to train the classifiers are explored.
The top six experiments for each classifier and feature extractors are shown in Figure \ref{fig:classifiertestresults}.

\subsubsection{K-Nearest Neighbour}
Comparing the weight function used to increase the impact of the distance, it is evident that when points are weighted by the inverse of their distance, it performs better by 1\%.
However, comparing the algorithm used to calculate the nearest neighbours, the K-dimension tree approach is 1.7\% slower compared to the brute force approach with the same accuracy.

\subsubsection{Naive Bayes}

The smoothing variable has minimal effect, with less than 0.5\% performance gain.

\subsubsection{Support Vector Machine}

Out of the four kernels that were tested, the polynomial kernel had the best performance over RBF, Linear and Sigmoid kernels with a performance gain of 0.7\%, 1.5\% and 2.7\% respectively.
The effect of the degree of the polynomial can be explored further since it is set to three for all the experiments.
In terms of the inference speed, the difference between the kernels is minimal.

Next, comparing the C (regularization) value, it can be observed that the C value greater than 100 is the ideal for polynomial, RBF and sigmoid compared to 0.01 for the linear kernel.
This is due to the generalization aspect of increasing the C value.

\subsubsection{Decision Tree}

The Gini Impurity criterion performs better than the Information Gain criterion.
It can achieve 16.5\% more accuracy while maintaining the same inference time.
Comparing the maximum depth, it is evident that at a depth of 10, the model is underfitting, not even able to obtain the excellent train accuracy.
However, the performance difference is minimal when the maximum depth is more than 50 .

The overall accuracy of the decision tree with model interpretability is the lowest out of all the classifiers.
So performing a model ensemble to form a random forest, creates a massive leap in performance.
However, the Information Gain criterion performs better than the Gini Impurity criterion, with a performance difference of 2\%.
The maximum depth of more than 50 has minimal effect on accuracy.
Nevertheless, the number of trees in the forest have a massive effect on performance.
A performance difference of 62\% is obtained by having 1000 trees compared to 2 trees.
Having different trees work on multiple subsets of the dataset along with majority voting allows more trees to perform and generalize better on indoor scene classification.

\subsubsection{Neural Network}

Analyzing the effect of a shallow and deep architecture, a 5.5\% decrease in performance in the deep architecture compared to shallow is evident.
This is mainly due to the deep architecture overfitting the data.
The training accuracy of the deep architecture is 0.9985 compared to 0.9868 of the shallow architecture.
The increase in parameters of the architecture causes the model to overfit, so having fewer parameters in the neural network helps to increase the performance.

Examining the effect of batch size, it is evident that there is little to no effect on the inference time and accuracy.
This is due to the VRAM of the GPU being able to handle both batch sizes easily.
This can be different if the batch size is larger than 64, but during inference, since the batch size is 1, this can be ignored.

Out of the three optimizers compared in this study, Adam optimizer performed the best followed by AdaMax and SGD.
The Adam optimizer was also reaching the optima at around 322 epochs on average compared to 325 and 336 of SGD and AdaMax respectively.
However, the AdaMax optimizer was able to achieve the highest accuracy among all of the experiments followed by Adam and SGD.

The learning rate also has a significant effect on how long it takes to converge.
The learning rate of 1e-03 can achieve similar performance as 1e-05 with far fewer epochs.
Having 1e-07 as the learning rate creates a significant dip in performance due to the model taking too long to converge, thus capping at the 500 epoch limit.
On the other hand, if the model is stuck in a local minimum, the learning rate is too small to come out of it.

Next, having the learning rate decay (lrDecay) increases performance by 0.67\%.
This is not a significant increase, but it can be understood that having the decay improves the performance.
On the deep architecture, the performance decreased by 2.7\%.
This is due to the optimizer being stuck in a local minimum, and the lrDecay reduced the chance of recovering from it.
So to improve this, trying different lrDecay values is recommended.
Also instead of stepping the lrDecay every ten epochs, other values can be used.

Early stopping and model checkpointing allowed for much better results compared to letting the neural network classifiers train for a fixed number of epochs.

\section{Conclusion} \label{conclusion}

It is evident that there exists a performance difference among the machine learning classifiers in indoor scene classification.
The neural network classifier has the best overall performance, with the main downside of a long and tedious training process.
On the other hand, the SVM classifier can be trained relatively fast and efficiently with the drawback of increase in inference time.
The decision tree classifier with its model interpretability has the worst performance. 
However, ensembling many trees to form a random forest increases the performance with the downside of losing model interpretability.
Naive Bayes is the simplest classifier with good performance, whereas the performance of the KNN classifier is poor.

Comparing SOTA low latency feature extractors to enormous feature extractor in indoor scene classification, it is evident that the performance drop was only 15\% where the speed increase was more than 300\%.
Therefore, MnasNet architecture would be suitable for real-time indoor scene recognition.

In conclusion, this study performs an extensive comparison into the performance of different machine learning algorithms and feature extractors, while introducing a simple novel low latency indoor scene classification system.

\section{Future Work}

This research study opens up multiple research paths, especially to identify the effect of finetuning on scene classification datasets such as Places365 \cite{zhou2017places}.
As shown by \cite{liu2019novel}, the effect of data augmentation is powerful on indoor scene classification. 
So, another path would be to introduce carefully curated data augmentation. 
With the benchmark and codebase publicly available, this study opens up the exploration for many future studies.

\ifCLASSOPTIONcaptionsoff
  \newpage
\fi



\bibliographystyle{IEEEtran}
\bibliography{references}

\onecolumn

\appendices
\section{Results}
\label{appx:results}
\begin{longtable}{p{0.08\linewidth}p{0.23\linewidth}p{0.075\linewidth}p{0.07\linewidth}p{0.07\linewidth}p{0.07\linewidth}p{0.07\linewidth}p{0.07\linewidth}}
  \toprule
      Classifier &                        Experiment Name &  Train Acc. &  Train F1 Score &  Train Time &  Test Acc. &  Test F1 Score &  Test Time \\
  \midrule
  \endhead
  \midrule
  \multicolumn{8}{r}{{Continued on next page}} \\
  \midrule
  \endfoot
  \bottomrule
  \endlastfoot
  \textbf{nn} &  \textbf{resnext101-shallow-64-adamax-1e-05-wd} &          \textbf{0.9868} &          \textbf{0.9868} &   \textbf{1290.2530} &         \textbf{0.7687} &         \textbf{0.7665} &     \textbf{0.0933} \\
  \textbf{svm} &                  \textbf{resnext101-poly-100.0} &          \textbf{0.9991} &          \textbf{0.9991} &    \textbf{530.3962} &         \textbf{0.7672} &        \textbf{0.7658} &     \textbf{0.1015} \\
 svm &                resnext101-poly-10000.0 &          0.9991 &          0.9991 &    530.4129 &         0.7672 &         0.7657 &     0.1014 \\
  nn &  resnext101-shallow-32-adamax-1e-05-wd &          0.9910 &          0.9910 &   1298.0639 &         0.7687 &         0.7657 &     0.0937 \\
  nn &    resnext101-shallow-64-adam-1e-05-wd &          0.9722 &          0.9722 &    615.4730 &         0.7679 &         0.7642 &     0.0934 \\
  nn &       resnext101-shallow-64-adam-1e-05 &          0.9968 &          0.9968 &    642.8912 &         0.7634 &         0.7617 &     0.0934 \\
 svm &                 resnext101-rbf-10000.0 &          0.9991 &          0.9991 &    530.9957 &         0.7612 &         0.7604 &     0.1023 \\
 svm &                   resnext101-rbf-100.0 &          0.9991 &          0.9991 &    531.0885 &         0.7612 &         0.7604 &     0.1023 \\
  nn &        resnext101-shallow-32-sgd-0.001 &          0.9634 &          0.9633 &    658.2545 &         0.7627 &         0.7596 &     0.0938 \\
  nn &     resnext101-shallow-64-adamax-1e-05 &          0.9457 &          0.9456 &    776.2874 &         0.7612 &         0.7591 &     0.0934 \\
 svm &                       resnext101-rbf-1 &          0.8629 &          0.8637 &    540.2303 &         0.7590 &         0.7586 &     0.1026 \\
 svm &                 resnext101-linear-0.01 &          0.9935 &          0.9935 &    524.2890 &         0.7552 &         0.7545 &     0.1016 \\
  nn &     resnext101-shallow-64-sgd-0.001-wd &          0.9649 &          0.9649 &    801.3545 &         0.7560 &         0.7543 &     0.0935 \\
  nn &     resnext101-shallow-32-adamax-1e-05 &          0.9854 &          0.9854 &    895.8429 &         0.7545 &         0.7521 &     0.0937 \\
  nn &    resnext101-shallow-32-adam-1e-05-wd &          0.9744 &          0.9745 &    607.6985 &         0.7515 &         0.7492 &     0.0936 \\
 svm &              resnext101-linear-10000.0 &          0.9991 &          0.9991 &    524.2180 &         0.7493 &         0.7485 &     0.1017 \\
 svm &                resnext101-linear-100.0 &          0.9991 &          0.9991 &    524.4187 &         0.7493 &         0.7485 &     0.1017 \\
 svm &                    resnext101-linear-1 &          0.9991 &          0.9991 &    524.1753 &         0.7493 &         0.7485 &     0.1017 \\
  nn &     resnext101-shallow-64-adamax-0.001 &          0.9382 &          0.9382 &    501.4016 &         0.7493 &         0.7484 &     0.0934 \\
  nn &  resnext101-shallow-64-adamax-0.001-wd &          0.9371 &          0.9371 &    501.3757 &         0.7500 &         0.7483 &     0.0934 \\
 svm &               resnext101-sigmoid-100.0 &          0.9985 &          0.9985 &    523.0545 &         0.7463 &         0.7469 &     0.1017 \\
 svm &             resnext101-sigmoid-10000.0 &          0.9991 &          0.9991 &    523.0373 &         0.7463 &         0.7468 &     0.1017 \\
  nn &        resnext101-shallow-64-sgd-0.001 &          0.9474 &          0.9473 &    723.9333 &         0.7478 &         0.7451 &     0.0935 \\
  nn &       resnext101-shallow-32-adam-1e-05 &          0.9976 &          0.9976 &    647.5013 &         0.7448 &         0.7422 &     0.0937 \\
  nn &     resnext101-shallow-32-adamax-0.001 &          0.9543 &          0.9543 &    504.9757 &         0.7425 &         0.7396 &     0.0936 \\
 svm &                   resnext101-sigmoid-1 &          0.7965 &          0.7980 &    544.6762 &         0.7388 &         0.7391 &     0.1027 \\
  nn &  resnext101-shallow-32-adamax-0.001-wd &          0.9981 &          0.9981 &    544.2368 &         0.7373 &         0.7361 &     0.0937 \\
  nn &       resnext101-shallow-64-adam-0.001 &          0.8668 &          0.8651 &    497.9940 &         0.7410 &         0.7352 &     0.0934 \\
  nn &     resnext101-shallow-32-sgd-0.001-wd &          0.9912 &          0.9912 &    799.2347 &         0.7328 &         0.7302 &     0.0937 \\
  nn &  mnasnet1\_0-shallow-64-adamax-0.001-wd &          0.9713 &          0.9712 &    123.7619 &         0.7299 &         0.7282 &     0.0231 \\
 svm &                 mnasnet1\_0-linear-0.01 &          0.9688 &          0.9689 &    137.3875 &         0.7276 &         0.7269 &     0.0280 \\
 svm &                   mnasnet1\_0-rbf-100.0 &          0.9991 &          0.9991 &    140.7178 &         0.7261 &         0.7249 &     0.0283 \\
 svm &                 mnasnet1\_0-rbf-10000.0 &          0.9991 &          0.9991 &    140.7831 &         0.7261 &         0.7249 &     0.0283 \\
  nn &    mnasnet1\_0-shallow-64-adam-1e-05-wd &          0.9965 &          0.9965 &    349.0961 &         0.7261 &         0.7238 &     0.0231 \\
  nn &          resnext101-deep-64-adam-1e-05 &          0.9985 &          0.9985 &   1048.7318 &         0.7261 &         0.7238 &     0.0934 \\
 svm &                mnasnet1\_0-poly-10000.0 &          0.9991 &          0.9991 &    150.7850 &         0.7231 &         0.7236 &     0.0279 \\
  nn &     mnasnet1\_0-shallow-32-adamax-1e-05 &          0.9731 &          0.9731 &    652.6965 &         0.7246 &         0.7216 &     0.0234 \\
\textbf{naive-bayes} &                       \textbf{resnext101-1e-09} &          \textbf{0.8254} &          \textbf{0.8262} &    \textbf{494.5109} &         \textbf{0.7187} &         \textbf{0.7204} &     \textbf{0.0926} \\
naive-bayes &                       resnext101-1e-06 &          0.8254 &          0.8262 &    494.5124 &         0.7187 &         0.7204 &     0.0926 \\
naive-bayes &                       resnext101-0.001 &          0.8228 &          0.8234 &    494.5123 &         0.7179 &         0.7194 &     0.0926 \\
 svm &              mnasnet1\_0-linear-10000.0 &          0.9991 &          0.9991 &    137.2108 &         0.7194 &         0.7191 &     0.0280 \\
 svm &                mnasnet1\_0-linear-100.0 &          0.9991 &          0.9991 &    137.2341 &         0.7194 &         0.7191 &     0.0280 \\
 svm &                    mnasnet1\_0-linear-1 &          0.9991 &          0.9991 &    137.2027 &         0.7194 &         0.7191 &     0.0280 \\
 svm &               mnasnet1\_0-sigmoid-100.0 &          0.9991 &          0.9991 &    137.4634 &         0.7187 &         0.7183 &     0.0280 \\
 svm &             mnasnet1\_0-sigmoid-10000.0 &          0.9991 &          0.9991 &    137.4820 &         0.7187 &         0.7183 &     0.0280 \\
 svm &                       mnasnet1\_0-rbf-1 &          0.8157 &          0.8164 &    151.9157 &         0.7172 &         0.7178 &     0.0287 \\
  nn &    resnext101-shallow-32-adam-0.001-wd &          0.9985 &          0.9985 &   1546.6470 &         0.7194 &         0.7175 &     0.0937 \\
  nn &    resnext101-shallow-64-adam-0.001-wd &          0.9849 &          0.9849 &    512.1320 &         0.7164 &         0.7163 &     0.0934 \\
 svm &                  mnasnet1\_0-poly-100.0 &          0.9537 &          0.9548 &    150.4568 &         0.7097 &         0.7155 &     0.0281 \\
  nn &    mnasnet1\_0-shallow-32-adam-1e-05-wd &          0.9840 &          0.9840 &    269.5522 &         0.7194 &         0.7153 &     0.0234 \\
  nn &       mnasnet1\_0-shallow-64-adam-1e-05 &          0.9806 &          0.9806 &    277.5261 &         0.7179 &         0.7153 &     0.0231 \\
  nn &     mnasnet1\_0-shallow-64-adamax-0.001 &          0.9989 &          0.9989 &    154.8707 &         0.7187 &         0.7151 &     0.0230 \\
  nn &     mnasnet1\_0-shallow-32-adamax-0.001 &          0.9319 &          0.9319 &    122.9749 &         0.7187 &         0.7148 &     0.0233 \\
  nn &  mnasnet1\_0-shallow-32-adamax-1e-05-wd &          0.9804 &          0.9804 &   1128.0963 &         0.7164 &         0.7132 &     0.0234 \\
  nn &  mnasnet1\_0-shallow-64-adamax-1e-05-wd &          0.9636 &          0.9635 &   1118.9030 &         0.7157 &         0.7129 &     0.0230 \\
  nn &     mnasnet1\_0-shallow-64-adamax-1e-05 &          0.9953 &          0.9953 &    739.8007 &         0.7157 &         0.7127 &     0.0230 \\
\textbf{random-forest} &            \textbf{resnext101-1000-entropy-100} &          \textbf{0.9991} &          \textbf{0.9991} &    \textbf{697.4053} &         \textbf{0.7194} &         \textbf{0.7103} &     \textbf{0.0926} \\
  nn &  mnasnet1\_0-shallow-32-adamax-0.001-wd &          0.9293 &          0.9292 &    123.0034 &         0.7112 &         0.7083 &     0.0234 \\
random-forest &             resnext101-1000-entropy-50 &          0.9991 &          0.9991 &    692.2018 &         0.7172 &         0.7079 &     0.0927 \\
random-forest &           resnext101-1000-entropy-None &          0.9991 &          0.9991 &    694.0079 &         0.7172 &         0.7076 &     0.0926 \\
  nn &     resnext101-deep-64-adamax-0.001-wd &          0.9981 &          0.9981 &   1154.6321 &         0.7082 &         0.7072 &     0.0934 \\
  nn &       mnasnet1\_0-shallow-32-adam-1e-05 &          0.9937 &          0.9937 &    281.0097 &         0.7060 &         0.7039 &     0.0234 \\
  nn &       resnext101-deep-64-adam-1e-05-wd &          0.9991 &          0.9991 &   1320.0112 &         0.7060 &         0.7039 &     0.0934 \\
random-forest &             resnext101-1000-entropy-10 &          0.9991 &          0.9991 &    691.2938 &         0.7134 &         0.7032 &     0.0927 \\
  nn &       resnext101-deep-64-adam-0.001-wd &          0.9976 &          0.9976 &   1332.0689 &         0.7052 &         0.7021 &     0.0934 \\
  nn &       resnext101-shallow-32-adam-0.001 &          0.9944 &          0.9944 &   1132.8347 &         0.7052 &         0.7021 &     0.0937 \\
  nn &    mnasnet1\_0-shallow-64-adam-0.001-wd &          0.8912 &          0.8908 &    119.7367 &         0.7037 &         0.7009 &     0.0231 \\
  nn &        resnext101-deep-64-adamax-1e-05 &          0.9772 &          0.9772 &   1205.9989 &         0.7022 &         0.6997 &     0.0934 \\
  nn &       resnext101-deep-32-adam-1e-05-wd &          0.9987 &          0.9987 &   1050.7437 &         0.7015 &         0.6986 &     0.0937 \\
  nn &     mnasnet1\_0-shallow-64-sgd-0.001-wd &          0.9746 &          0.9746 &    686.2455 &         0.7007 &         0.6963 &     0.0231 \\
  nn &        resnext101-deep-64-adamax-0.001 &          0.9961 &          0.9961 &    777.3307 &         0.6985 &         0.6955 &     0.0934 \\
  nn &       mnasnet1\_0-shallow-64-adam-0.001 &          0.8942 &          0.8938 &    119.7749 &         0.6940 &         0.6939 &     0.0231 \\
 svm &                   mnasnet1\_0-sigmoid-1 &          0.7610 &          0.7626 &    157.3438 &         0.6903 &         0.6934 &     0.0286 \\
  nn &          resnext101-deep-32-adam-1e-05 &          0.9972 &          0.9972 &    884.1581 &         0.6948 &         0.6926 &     0.0937 \\
random-forest &              resnext101-1000-gini-None &          0.9991 &          0.9991 &    552.9757 &         0.7030 &         0.6925 &     0.0926 \\
random-forest &                resnext101-1000-gini-50 &          0.9991 &          0.9991 &    556.7797 &         0.7022 &         0.6906 &     0.0927 \\
  nn &           resnext101-deep-32-sgd-0.001 &          0.9976 &          0.9976 &   1257.7591 &         0.6903 &         0.6901 &     0.0938 \\
  nn &        mnasnet1\_0-shallow-32-sgd-0.001 &          0.9994 &          0.9994 &    800.6678 &         0.6925 &         0.6899 &     0.0235 \\
  nn &        resnext101-deep-32-adamax-0.001 &          0.9953 &          0.9953 &   1441.8088 &         0.6925 &         0.6894 &     0.0937 \\
 svm &                      resnext101-poly-1 &          0.7541 &          0.7601 &    566.6230 &         0.6858 &         0.6894 &     0.1027 \\
  nn &        resnext101-deep-32-adamax-1e-05 &          0.9924 &          0.9924 &   1451.3782 &         0.6866 &         0.6846 &     0.0937 \\
  nn &        mnasnet1\_0-shallow-64-sgd-0.001 &          0.9938 &          0.9938 &    604.3969 &         0.6881 &         0.6843 &     0.0232 \\
  nn &     mnasnet1\_0-shallow-32-sgd-0.001-wd &          0.9681 &          0.9680 &    365.6768 &         0.6836 &         0.6835 &     0.0235 \\
random-forest &               resnext101-1000-gini-100 &          0.9991 &          0.9991 &    554.3802 &         0.6933 &         0.6828 &     0.0927 \\
  nn &     resnext101-deep-32-adamax-0.001-wd &          0.9985 &          0.9985 &   1204.0463 &         0.6866 &         0.6828 &     0.0937 \\
  nn &       resnext101-deep-32-adam-0.001-wd &          0.9950 &          0.9950 &   1352.3992 &         0.6866 &         0.6826 &     0.0938 \\
random-forest &               mnasnet1\_0-1000-gini-100 &          0.9991 &          0.9991 &    146.6666 &         0.6903 &         0.6802 &     0.0221 \\
random-forest &            mnasnet1\_0-1000-entropy-100 &          0.9991 &          0.9991 &    249.5366 &         0.6881 &         0.6801 &     0.0221 \\
random-forest &                mnasnet1\_0-1000-gini-50 &          0.9991 &          0.9991 &    147.3597 &         0.6873 &         0.6759 &     0.0221 \\
random-forest &           mnasnet1\_0-1000-entropy-None &          0.9991 &          0.9991 &    250.7377 &         0.6843 &         0.6752 &     0.0221 \\
random-forest &             mnasnet1\_0-1000-entropy-50 &          0.9991 &          0.9991 &    248.7456 &         0.6851 &         0.6729 &     0.0221 \\
random-forest &              mnasnet1\_0-1000-gini-None &          0.9991 &          0.9991 &    144.1507 &         0.6813 &         0.6724 &     0.0221 \\
  nn &       mnasnet1\_0-shallow-32-adam-0.001 &          0.9931 &          0.9931 &    273.2562 &         0.6776 &         0.6724 &     0.0234 \\
  nn &    mnasnet1\_0-shallow-32-adam-0.001-wd &          0.9994 &          0.9994 &   1461.3441 &         0.6724 &         0.6685 &     0.0234 \\
  nn &          resnext101-deep-64-adam-0.001 &          0.9912 &          0.9912 &   1397.2174 &         0.6731 &         0.6683 &     0.0934 \\
  nn &          resnext101-deep-32-adam-0.001 &          0.9826 &          0.9827 &   1380.8360 &         0.6724 &         0.6670 &     0.0937 \\
  nn &       mnasnet1\_0-deep-64-adam-0.001-wd &          0.9989 &          0.9989 &   1313.4450 &         0.6687 &         0.6668 &     0.0231 \\
  nn &        resnext101-deep-64-sgd-0.001-wd &          0.9739 &          0.9739 &   1319.4682 &         0.6679 &         0.6653 &     0.0935 \\
  nn &        resnext101-deep-32-sgd-0.001-wd &          0.9924 &          0.9923 &   1303.3129 &         0.6664 &         0.6644 &     0.0938 \\
  nn &     mnasnet1\_0-deep-32-adamax-0.001-wd &          0.9989 &          0.9989 &   1356.6482 &         0.6642 &         0.6618 &     0.0234 \\
  nn &           resnext101-deep-64-sgd-0.001 &          0.9276 &          0.9273 &    796.6236 &         0.6657 &         0.6613 &     0.0935 \\
  nn &     mnasnet1\_0-deep-64-adamax-0.001-wd &          0.9991 &          0.9991 &    825.5003 &         0.6657 &         0.6612 &     0.0231 \\
random-forest &                resnext101-100-gini-100 &          0.9991 &          0.9991 &    500.4926 &         0.6672 &         0.6562 &     0.0922 \\
random-forest &             mnasnet1\_0-1000-entropy-10 &          0.9991 &          0.9991 &    249.0943 &         0.6679 &         0.6559 &     0.0221 \\
naive-bayes &                       mnasnet1\_0-0.001 &          0.8409 &          0.8413 &    115.6235 &         0.6515 &         0.6542 &     0.0218 \\
  nn &          mnasnet1\_0-deep-64-adam-0.001 &          0.9933 &          0.9933 &    696.0314 &         0.6582 &         0.6519 &     0.0231 \\
random-forest &             resnext101-100-entropy-100 &          0.9991 &          0.9991 &    514.8122 &         0.6612 &         0.6518 &     0.0922 \\
random-forest &            resnext101-100-entropy-None &          0.9991 &          0.9991 &    515.0193 &         0.6590 &         0.6499 &     0.0922 \\
naive-bayes &                       mnasnet1\_0-1e-06 &          0.8446 &          0.8450 &    115.6239 &         0.6448 &         0.6482 &     0.0218 \\
naive-bayes &                       mnasnet1\_0-1e-09 &          0.8446 &          0.8450 &    115.6242 &         0.6448 &         0.6482 &     0.0218 \\
  nn &        mnasnet1\_0-deep-32-adamax-0.001 &          0.9970 &          0.9970 &    797.6790 &         0.6493 &         0.6466 &     0.0234 \\
 \textbf{knn} &              \textbf{resnext101-brute-distance} &          \textbf{0.9991} &          \textbf{0.9991} &    \textbf{494.4501} &         \textbf{0.6575} &         \textbf{0.6462} &     \textbf{0.0923} \\
 knn &          resnext101-ball\_tree-distance &          0.9991 &          0.9991 &    494.9604 &         0.6575 &         0.6462 &     0.0932 \\
 knn &            resnext101-kd\_tree-distance &          0.9991 &          0.9991 &    494.9162 &         0.6575 &         0.6462 &     0.0940 \\
random-forest &              resnext101-100-entropy-10 &          0.9989 &          0.9989 &    514.0149 &         0.6560 &         0.6460 &     0.0922 \\
random-forest &                 resnext101-100-gini-50 &          0.9991 &          0.9991 &    500.3930 &         0.6552 &         0.6454 &     0.0922 \\
  nn &          mnasnet1\_0-deep-32-adam-1e-05 &          0.9991 &          0.9991 &   1097.4465 &         0.6485 &         0.6446 &     0.0235 \\
random-forest &               resnext101-100-gini-None &          0.9991 &          0.9991 &    500.2921 &         0.6515 &         0.6439 &     0.0922 \\
  nn &     resnext101-deep-64-adamax-1e-05-wd &          0.8647 &          0.8617 &   1333.3698 &         0.6545 &         0.6436 &     0.0934 \\
  nn &        mnasnet1\_0-deep-64-adamax-0.001 &          0.9981 &          0.9981 &    829.4137 &         0.6463 &         0.6417 &     0.0231 \\
  nn &       mnasnet1\_0-deep-64-adam-1e-05-wd &          0.9979 &          0.9979 &    998.7173 &         0.6455 &         0.6404 &     0.0231 \\
 knn &               resnext101-brute-uniform &          0.6647 &          0.6594 &    494.4509 &         0.6507 &         0.6399 &     0.0923 \\
 knn &           resnext101-ball\_tree-uniform &          0.6647 &          0.6594 &    495.0360 &         0.6507 &         0.6399 &     0.0932 \\
 knn &             resnext101-kd\_tree-uniform &          0.6647 &          0.6594 &    494.9103 &         0.6507 &         0.6399 &     0.0940 \\
random-forest &              resnext101-100-entropy-50 &          0.9991 &          0.9991 &    515.1205 &         0.6463 &         0.6357 &     0.0922 \\
  nn &          mnasnet1\_0-deep-64-adam-1e-05 &          0.9991 &          0.9991 &   1052.5357 &         0.6381 &         0.6355 &     0.0231 \\
  nn &     resnext101-deep-32-adamax-1e-05-wd &          0.8696 &          0.8667 &   1524.0322 &         0.6440 &         0.6338 &     0.0936 \\
  nn &       mnasnet1\_0-deep-32-adam-1e-05-wd &          0.9981 &          0.9981 &   1155.9426 &         0.6351 &         0.6319 &     0.0234 \\
  nn &       mnasnet1\_0-deep-32-adam-0.001-wd &          0.9966 &          0.9966 &   1103.7082 &         0.6306 &         0.6272 &     0.0235 \\
  nn &        mnasnet1\_0-deep-64-adamax-1e-05 &          0.9720 &          0.9718 &   1079.1000 &         0.6284 &         0.6247 &     0.0231 \\
  nn &          mnasnet1\_0-deep-32-adam-0.001 &          0.9899 &          0.9899 &   1255.6537 &         0.6291 &         0.6242 &     0.0234 \\
  nn &           mnasnet1\_0-deep-64-sgd-0.001 &          0.9968 &          0.9968 &   1153.4391 &         0.6231 &         0.6241 &     0.0232 \\
random-forest &               mnasnet1\_0-100-gini-None &          0.9991 &          0.9991 &    118.9210 &         0.6224 &         0.6138 &     0.0217 \\
  nn &        mnasnet1\_0-deep-32-adamax-1e-05 &          0.9509 &          0.9508 &   1146.0586 &         0.6194 &         0.6115 &     0.0234 \\
random-forest &                mnasnet1\_0-1000-gini-10 &          0.9957 &          0.9957 &    132.5350 &         0.6299 &         0.6099 &     0.0221 \\
  nn &           mnasnet1\_0-deep-32-sgd-0.001 &          0.9920 &          0.9920 &    725.7567 &         0.6060 &         0.6062 &     0.0235 \\
 knn &          mnasnet1\_0-ball\_tree-distance &          0.9991 &          0.9991 &    115.8433 &         0.6037 &         0.6059 &     0.0223 \\
 knn &            mnasnet1\_0-kd\_tree-distance &          0.9991 &          0.9991 &    115.8088 &         0.6037 &         0.6059 &     0.0229 \\
 knn &              mnasnet1\_0-brute-distance &          0.9991 &          0.9991 &    115.5803 &         0.6037 &         0.6058 &     0.0217 \\
random-forest &                mnasnet1\_0-100-gini-100 &          0.9991 &          0.9991 &    118.6196 &         0.6075 &         0.5988 &     0.0217 \\
 knn &               mnasnet1\_0-brute-uniform &          0.6241 &          0.6237 &    115.5802 &         0.5881 &         0.5896 &     0.0217 \\
 knn &           mnasnet1\_0-ball\_tree-uniform &          0.6241 &          0.6237 &    115.8442 &         0.5881 &         0.5896 &     0.0223 \\
 knn &             mnasnet1\_0-kd\_tree-uniform &          0.6241 &          0.6237 &    115.7899 &         0.5881 &         0.5896 &     0.0229 \\
random-forest &                 mnasnet1\_0-100-gini-50 &          0.9991 &          0.9991 &    118.5203 &         0.5925 &         0.5825 &     0.0217 \\
  nn &        mnasnet1\_0-deep-64-sgd-0.001-wd &          0.9416 &          0.9415 &    867.3524 &         0.5843 &         0.5804 &     0.0232 \\
  nn &        mnasnet1\_0-deep-32-sgd-0.001-wd &          0.9853 &          0.9853 &    971.4710 &         0.5828 &         0.5777 &     0.0235 \\
random-forest &                resnext101-1000-gini-10 &          0.9547 &          0.9549 &    520.8756 &         0.5955 &         0.5694 &     0.0927 \\
random-forest &              mnasnet1\_0-100-entropy-10 &          0.9991 &          0.9991 &    128.9397 &         0.5806 &         0.5646 &     0.0217 \\
random-forest &            mnasnet1\_0-100-entropy-None &          0.9991 &          0.9991 &    129.1390 &         0.5739 &         0.5600 &     0.0217 \\
random-forest &              mnasnet1\_0-100-entropy-50 &          0.9991 &          0.9991 &    129.2369 &         0.5731 &         0.5595 &     0.0217 \\
random-forest &             mnasnet1\_0-100-entropy-100 &          0.9991 &          0.9991 &    129.2396 &         0.5657 &         0.5535 &     0.0217 \\
  nn &     mnasnet1\_0-deep-32-adamax-1e-05-wd &          0.7886 &          0.7749 &   1435.5942 &         0.5694 &         0.5487 &     0.0234 \\
 svm &               resnext101-linear-0.0001 &          0.5950 &          0.5845 &    581.5421 &         0.5522 &         0.5418 &     0.1033 \\
  nn &     mnasnet1\_0-deep-64-adamax-1e-05-wd &          0.7511 &          0.7401 &   1228.4915 &         0.5627 &         0.5400 &     0.0231 \\
random-forest &                 mnasnet1\_0-100-gini-10 &          0.9832 &          0.9832 &    117.4183 &         0.5515 &         0.5263 &     0.0217 \\
  nn &       resnext101-shallow-32-adam-1e-07 &          0.6343 &          0.6137 &   1604.6319 &         0.5545 &         0.5180 &     0.0937 \\
random-forest &                 resnext101-100-gini-10 &          0.8888 &          0.8875 &    497.2872 &         0.5254 &         0.4933 &     0.0922 \\
  nn &       resnext101-shallow-64-adam-1e-07 &          0.5840 &          0.5555 &   1362.7278 &         0.5187 &         0.4804 &     0.0934 \\
random-forest &                resnext101-10-gini-None &          0.9966 &          0.9966 &    495.3587 &         0.4813 &         0.4788 &     0.0922 \\
random-forest &                 resnext101-10-gini-100 &          0.9972 &          0.9972 &    495.3571 &         0.4784 &         0.4746 &     0.0922 \\
random-forest &                  resnext101-10-gini-50 &          0.9966 &          0.9966 &    495.5603 &         0.4724 &         0.4670 &     0.0922 \\
  nn &        resnext101-shallow-32-sgd-1e-05 &          0.5715 &          0.5375 &   1603.1219 &         0.5052 &         0.4577 &     0.0937 \\
  nn &       mnasnet1\_0-shallow-32-adam-1e-07 &          0.5405 &          0.5169 &   1456.5810 &         0.4597 &         0.4286 &     0.0234 \\
random-forest &               resnext101-10-entropy-10 &          0.9806 &          0.9806 &    496.9616 &         0.4306 &         0.4190 &     0.0922 \\
random-forest &              resnext101-10-entropy-100 &          0.9961 &          0.9961 &    497.0622 &         0.4157 &         0.4175 &     0.0922 \\
random-forest &             resnext101-10-entropy-None &          0.9976 &          0.9976 &    497.1645 &         0.4201 &         0.4173 &     0.0922 \\
random-forest &               resnext101-10-entropy-50 &          0.9968 &          0.9968 &    497.1622 &         0.4104 &         0.4104 &     0.0922 \\
  nn &    resnext101-shallow-32-adam-1e-07-wd &          0.4785 &          0.4471 &   1605.7025 &         0.4194 &         0.3882 &     0.0937 \\
\textbf{decision-tree} &                   \textbf{resnext101-gini-None} &          \textbf{0.9991} &          \textbf{0.9991} &    \textbf{538.5621} &         \textbf{0.3784} &         \textbf{0.3790} &     \textbf{0.0922} \\
decision-tree &                     resnext101-gini-50 &          0.9493 &          0.9554 &    537.6535 &         0.3791 &         0.3777 &     0.0922 \\
decision-tree &                    resnext101-gini-100 &          0.9991 &          0.9991 &    538.5325 &         0.3776 &         0.3769 &     0.0922 \\
random-forest &                  resnext101-10-gini-10 &          0.5882 &          0.5784 &    494.8573 &         0.3858 &         0.3623 &     0.0922 \\
  nn &        mnasnet1\_0-shallow-32-sgd-1e-05 &          0.4651 &          0.4397 &   1411.0009 &         0.3896 &         0.3569 &     0.0235 \\
  nn &        resnext101-shallow-64-sgd-1e-05 &          0.4547 &          0.4191 &   1388.2858 &         0.3903 &         0.3523 &     0.0935 \\
  nn &       mnasnet1\_0-shallow-64-adam-1e-07 &          0.4666 &          0.4433 &   1239.2570 &         0.3739 &         0.3469 &     0.0231 \\
random-forest &                  mnasnet1\_0-10-gini-50 &          0.9974 &          0.9974 &    116.0882 &         0.3351 &         0.3371 &     0.0217 \\
  nn &     resnext101-shallow-32-sgd-1e-05-wd &          0.4174 &          0.3776 &   1574.2232 &         0.3739 &         0.3314 &     0.0937 \\
random-forest &                 mnasnet1\_0-10-gini-100 &          0.9961 &          0.9961 &    116.0896 &         0.3306 &         0.3313 &     0.0217 \\
  nn &    resnext101-shallow-64-adam-1e-07-wd &          0.3864 &          0.3544 &   1300.7596 &         0.3560 &         0.3264 &     0.0934 \\
decision-tree &                resnext101-entropy-None &          0.9991 &          0.9991 &    619.9747 &         0.3157 &         0.3147 &     0.0922 \\
random-forest &                  mnasnet1\_0-10-gini-10 &          0.6993 &          0.6914 &    115.8886 &         0.3291 &         0.3130 &     0.0217 \\
decision-tree &                 resnext101-entropy-100 &          0.9991 &          0.9991 &    621.3826 &         0.3097 &         0.3081 &     0.0922 \\
random-forest &               mnasnet1\_0-10-entropy-10 &          0.9793 &          0.9793 &    117.3922 &         0.3157 &         0.3054 &     0.0217 \\
decision-tree &                  resnext101-entropy-10 &          0.7440 &          0.7451 &    619.0062 &         0.3045 &         0.3020 &     0.0922 \\
random-forest &                mnasnet1\_0-10-gini-None &          0.9965 &          0.9965 &    116.0889 &         0.3045 &         0.2984 &     0.0217 \\
decision-tree &                  resnext101-entropy-50 &          0.9991 &          0.9991 &    621.8744 &         0.2985 &         0.2965 &     0.0922 \\
random-forest &               mnasnet1\_0-10-entropy-50 &          0.9970 &          0.9970 &    117.3913 &         0.2896 &         0.2906 &     0.0217 \\
random-forest &              mnasnet1\_0-10-entropy-100 &          0.9963 &          0.9963 &    117.3825 &         0.2866 &         0.2827 &     0.0217 \\
random-forest &             mnasnet1\_0-10-entropy-None &          0.9976 &          0.9976 &    117.3932 &         0.2672 &         0.2673 &     0.0217 \\
  nn &     resnext101-shallow-32-adamax-1e-07 &          0.3297 &          0.3042 &   1498.7338 &         0.2866 &         0.2613 &     0.0939 \\
random-forest &                   resnext101-2-gini-50 &          0.7226 &          0.7199 &    495.1562 &         0.2709 &         0.2609 &     0.0922 \\
random-forest &                 resnext101-2-gini-None &          0.7228 &          0.7201 &    495.0546 &         0.2716 &         0.2590 &     0.0922 \\
decision-tree &                    mnasnet1\_0-gini-100 &          0.9991 &          0.9991 &    135.8964 &         0.2448 &         0.2453 &     0.0216 \\
random-forest &                  resnext101-2-gini-100 &          0.7254 &          0.7227 &    495.0547 &         0.2672 &         0.2445 &     0.0922 \\
random-forest &                resnext101-2-entropy-10 &          0.6265 &          0.6234 &    496.3626 &         0.2545 &         0.2430 &     0.0922 \\
  nn &           resnext101-deep-32-sgd-1e-05 &          0.2890 &          0.2563 &   1671.6346 &         0.2672 &         0.2404 &     0.0938 \\
decision-tree &                   mnasnet1\_0-gini-None &          0.9991 &          0.9991 &    135.6716 &         0.2403 &         0.2362 &     0.0216 \\
decision-tree &                     mnasnet1\_0-gini-50 &          0.9696 &          0.9741 &    135.5417 &         0.2328 &         0.2327 &     0.0216 \\
  nn &    mnasnet1\_0-shallow-32-adam-1e-07-wd &          0.2937 &          0.2754 &   1451.7953 &         0.2485 &         0.2310 &     0.0234 \\
random-forest &                   resnext101-2-gini-10 &          0.3401 &          0.3600 &    494.7537 &         0.2306 &         0.2278 &     0.0922 \\
  nn &        mnasnet1\_0-shallow-64-sgd-1e-05 &          0.2812 &          0.2572 &   1305.2016 &         0.2530 &         0.2263 &     0.0231 \\
  nn &     resnext101-shallow-64-adamax-1e-07 &          0.2741 &          0.2492 &   1308.5121 &         0.2396 &         0.2120 &     0.0934 \\
decision-tree &                  mnasnet1\_0-entropy-10 &          0.6864 &          0.6876 &    185.3211 &         0.2052 &         0.2056 &     0.0216 \\
decision-tree &                  mnasnet1\_0-entropy-50 &          0.9991 &          0.9991 &    186.6609 &         0.2045 &         0.2052 &     0.0216 \\
  nn &     resnext101-shallow-64-sgd-1e-05-wd &          0.2580 &          0.2358 &   1431.9030 &         0.2269 &         0.2047 &     0.0935 \\
random-forest &                resnext101-2-entropy-50 &          0.7088 &          0.7049 &    496.4548 &         0.2149 &         0.2046 &     0.0922 \\
decision-tree &                mnasnet1\_0-entropy-None &          0.9991 &          0.9991 &    186.7215 &         0.2052 &         0.2040 &     0.0216 \\
decision-tree &                 mnasnet1\_0-entropy-100 &          0.9991 &          0.9991 &    186.8131 &         0.2015 &         0.2028 &     0.0216 \\
random-forest &               resnext101-2-entropy-100 &          0.7063 &          0.7032 &    496.4634 &         0.2045 &         0.2010 &     0.0922 \\
random-forest &              resnext101-2-entropy-None &          0.7132 &          0.7111 &    496.3619 &         0.2067 &         0.1968 &     0.0922 \\
  nn &     mnasnet1\_0-shallow-32-sgd-1e-05-wd &          0.2541 &          0.2349 &   1422.3452 &         0.2119 &         0.1899 &     0.0235 \\
random-forest &                   mnasnet1\_0-2-gini-10 &          0.3771 &          0.3852 &    115.7830 &         0.1821 &         0.1795 &     0.0217 \\
random-forest &                mnasnet1\_0-2-entropy-10 &          0.6075 &          0.6063 &    116.7892 &         0.1739 &         0.1732 &     0.0217 \\
decision-tree &                     resnext101-gini-10 &          0.2416 &          0.2639 &    511.9497 &         0.1701 &         0.1726 &     0.0922 \\
  nn &          resnext101-deep-32-adam-1e-07 &          0.2188 &          0.2012 &   1522.4011 &         0.1761 &         0.1562 &     0.0937 \\
  nn &    mnasnet1\_0-shallow-64-adam-1e-07-wd &          0.2065 &          0.1953 &   1266.9201 &         0.1634 &         0.1558 &     0.0231 \\
random-forest &                  mnasnet1\_0-2-gini-100 &          0.6925 &          0.6873 &    115.9856 &         0.1672 &         0.1548 &     0.0217 \\
  nn &          resnext101-deep-64-adam-1e-07 &          0.1823 &          0.1694 &   1404.9960 &         0.1590 &         0.1448 &     0.0934 \\
random-forest &                   mnasnet1\_0-2-gini-50 &          0.6806 &          0.6767 &    115.8844 &         0.1522 &         0.1444 &     0.0217 \\
decision-tree &                     mnasnet1\_0-gini-10 &          0.2687 &          0.3021 &    124.7518 &         0.1485 &         0.1410 &     0.0216 \\
random-forest &                 mnasnet1\_0-2-gini-None &          0.6903 &          0.6854 &    115.9854 &         0.1515 &         0.1406 &     0.0217 \\
  nn &           resnext101-deep-64-sgd-1e-05 &          0.1715 &          0.1517 &   1419.4391 &         0.1545 &         0.1402 &     0.0935 \\
random-forest &              mnasnet1\_0-2-entropy-None &          0.6853 &          0.6806 &    116.8872 &         0.1493 &         0.1397 &     0.0217 \\
random-forest &               mnasnet1\_0-2-entropy-100 &          0.6808 &          0.6779 &    116.8906 &         0.1470 &         0.1339 &     0.0217 \\
  nn &           mnasnet1\_0-deep-32-sgd-1e-05 &          0.1649 &          0.1504 &   1509.5415 &         0.1455 &         0.1268 &     0.0235 \\
  nn &     mnasnet1\_0-shallow-32-adamax-1e-07 &          0.1528 &          0.1390 &   1325.1360 &         0.1388 &         0.1263 &     0.0233 \\
random-forest &                mnasnet1\_0-2-entropy-50 &          0.6761 &          0.6728 &    116.8896 &         0.1284 &         0.1233 &     0.0217 \\
  nn &        resnext101-deep-32-sgd-1e-05-wd &          0.1358 &          0.1226 &   1616.5724 &         0.1209 &         0.1109 &     0.0938 \\
  nn &  resnext101-shallow-32-adamax-1e-07-wd &          0.1388 &          0.1286 &   1542.3052 &         0.1104 &         0.1010 &     0.0936 \\
  nn &     mnasnet1\_0-shallow-64-adamax-1e-07 &          0.1093 &          0.1020 &   1230.3000 &         0.0985 &         0.0921 &     0.0230 \\
  nn &     mnasnet1\_0-shallow-64-sgd-1e-05-wd &          0.1112 &          0.1032 &   1156.6570 &         0.0940 &         0.0878 &     0.0232 \\
  nn &          mnasnet1\_0-deep-32-adam-1e-07 &          0.1149 &          0.1046 &   1325.7920 &         0.0925 &         0.0871 &     0.0234 \\
  nn &          mnasnet1\_0-deep-64-adam-1e-07 &          0.0937 &          0.0880 &   1285.1960 &         0.0813 &         0.0745 &     0.0231 \\
  nn &  resnext101-shallow-64-adamax-1e-07-wd &          0.0942 &          0.0910 &   1301.9637 &         0.0784 &         0.0743 &     0.0933 \\
  nn &       resnext101-deep-32-adam-1e-07-wd &          0.0858 &          0.0797 &   1618.0272 &         0.0806 &         0.0726 &     0.0937 \\
  nn &           mnasnet1\_0-deep-64-sgd-1e-05 &          0.0701 &          0.0643 &   1347.3848 &         0.0672 &         0.0582 &     0.0232 \\
  nn &  mnasnet1\_0-shallow-32-adamax-1e-07-wd &          0.0576 &          0.0547 &   1299.5256 &         0.0597 &         0.0557 &     0.0233 \\
  nn &       resnext101-deep-64-adam-1e-07-wd &          0.0728 &          0.0670 &   1342.2672 &         0.0597 &         0.0532 &     0.0934 \\
  nn &        mnasnet1\_0-deep-32-sgd-1e-05-wd &          0.0603 &          0.0566 &   1505.4828 &         0.0537 &         0.0512 &     0.0235 \\
  nn &        resnext101-deep-64-sgd-1e-05-wd &          0.0593 &          0.0554 &   1232.4259 &         0.0537 &         0.0493 &     0.0935 \\
  nn &  mnasnet1\_0-shallow-64-adamax-1e-07-wd &          0.0461 &          0.0435 &   1106.0471 &         0.0433 &         0.0398 &     0.0230 \\
  nn &        resnext101-deep-32-adamax-1e-07 &          0.0534 &          0.0491 &   1424.1569 &         0.0425 &         0.0393 &     0.0937 \\
  nn &       mnasnet1\_0-deep-32-adam-1e-07-wd &          0.0369 &          0.0349 &   1082.5452 &         0.0381 &         0.0365 &     0.0234 \\
 svm &               mnasnet1\_0-linear-0.0001 &          0.0500 &          0.0492 &    181.0784 &         0.0381 &         0.0345 &     0.0287 \\
  nn &        resnext101-deep-64-adamax-1e-07 &          0.0386 &          0.0357 &   1251.9877 &         0.0366 &         0.0326 &     0.0934 \\
  nn &     resnext101-deep-32-adamax-1e-07-wd &          0.0271 &          0.0255 &   1233.7407 &         0.0321 &         0.0300 &     0.0936 \\
  nn &        mnasnet1\_0-deep-64-adamax-1e-07 &          0.0257 &          0.0239 &   1118.7943 &         0.0291 &         0.0282 &     0.0230 \\
  nn &        mnasnet1\_0-deep-64-sgd-1e-05-wd &          0.0338 &          0.0324 &    933.3751 &         0.0299 &         0.0272 &     0.0232 \\
  nn &       mnasnet1\_0-deep-64-adam-1e-07-wd &          0.0282 &          0.0265 &    722.2695 &         0.0261 &         0.0251 &     0.0231 \\
  nn &        mnasnet1\_0-deep-32-adamax-1e-07 &          0.0256 &          0.0246 &   1265.8049 &         0.0216 &         0.0213 &     0.0233 \\
  nn &     resnext101-shallow-64-sgd-1e-07-wd &          0.0183 &          0.0176 &   1077.6604 &         0.0194 &         0.0192 &     0.0934 \\
  nn &           mnasnet1\_0-deep-64-sgd-1e-07 &          0.0159 &          0.0156 &   1113.0549 &         0.0187 &         0.0182 &     0.0231 \\
  nn &     mnasnet1\_0-deep-64-adamax-1e-07-wd &          0.0166 &          0.0155 &   1114.8170 &         0.0187 &         0.0181 &     0.0230 \\
  nn &     resnext101-deep-64-adamax-1e-07-wd &          0.0248 &          0.0228 &   1025.0829 &         0.0201 &         0.0179 &     0.0933 \\
  nn &     resnext101-shallow-32-sgd-1e-07-wd &          0.0179 &          0.0174 &    659.7547 &         0.0179 &         0.0177 &     0.0938 \\
  nn &     mnasnet1\_0-shallow-64-sgd-1e-07-wd &          0.0159 &          0.0156 &    303.7959 &         0.0164 &         0.0168 &     0.0231 \\
  nn &     mnasnet1\_0-shallow-32-sgd-1e-07-wd &          0.0159 &          0.0163 &    507.9781 &         0.0172 &         0.0167 &     0.0234 \\
  nn &           resnext101-deep-32-sgd-1e-07 &          0.0190 &          0.0176 &   1669.1323 &         0.0179 &         0.0165 &     0.0938 \\
  nn &           resnext101-deep-64-sgd-1e-07 &          0.0146 &          0.0147 &    663.6759 &         0.0164 &         0.0162 &     0.0935 \\
  nn &        mnasnet1\_0-shallow-32-sgd-1e-07 &          0.0138 &          0.0132 &    994.2681 &         0.0179 &         0.0161 &     0.0235 \\
  nn &        resnext101-deep-64-sgd-1e-07-wd &          0.0114 &          0.0104 &    975.8959 &         0.0179 &         0.0159 &     0.0934 \\
  nn &     mnasnet1\_0-deep-32-adamax-1e-07-wd &          0.0177 &          0.0170 &    943.2605 &         0.0149 &         0.0146 &     0.0234 \\
  nn &        resnext101-shallow-32-sgd-1e-07 &          0.0174 &          0.0170 &   1529.6792 &         0.0149 &         0.0145 &     0.0937 \\
  nn &        mnasnet1\_0-shallow-64-sgd-1e-07 &          0.0123 &          0.0127 &    605.4862 &         0.0149 &         0.0140 &     0.0231 \\
  nn &        mnasnet1\_0-deep-32-sgd-1e-07-wd &          0.0131 &          0.0127 &    287.0596 &         0.0134 &         0.0124 &     0.0235 \\
  nn &        resnext101-deep-32-sgd-1e-07-wd &          0.0159 &          0.0150 &   1612.6127 &         0.0134 &         0.0120 &     0.0937 \\
  nn &        resnext101-shallow-64-sgd-1e-07 &          0.0136 &          0.0139 &   1245.5499 &         0.0112 &         0.0108 &     0.0934 \\
  nn &           mnasnet1\_0-deep-32-sgd-1e-07 &          0.0177 &          0.0163 &    180.1698 &         0.0112 &         0.0107 &     0.0235 \\
  nn &        mnasnet1\_0-deep-64-sgd-1e-07-wd &          0.0159 &          0.0149 &    882.6261 &         0.0097 &         0.0075 &     0.0231 \\
 svm &                      mnasnet1\_0-poly-1 &          0.0155 &          0.0005 &    181.6799 &         0.0149 &         0.0045 &     0.0287 \\
 svm &                   mnasnet1\_0-poly-0.01 &          0.0155 &          0.0005 &    181.1806 &         0.0127 &         0.0003 &     0.0287 \\
 svm &                mnasnet1\_0-sigmoid-0.01 &          0.0155 &          0.0005 &    182.6480 &         0.0127 &         0.0003 &     0.0287 \\
 svm &                 mnasnet1\_0-poly-0.0001 &          0.0155 &          0.0005 &    181.1163 &         0.0127 &         0.0003 &     0.0287 \\
 svm &              mnasnet1\_0-sigmoid-0.0001 &          0.0155 &          0.0005 &    182.4645 &         0.0127 &         0.0003 &     0.0287 \\
 svm &                    mnasnet1\_0-rbf-0.01 &          0.0155 &          0.0005 &    183.7749 &         0.0127 &         0.0003 &     0.0289 \\
 svm &                  mnasnet1\_0-rbf-0.0001 &          0.0155 &          0.0005 &    183.5261 &         0.0127 &         0.0003 &     0.0290 \\
 svm &                   resnext101-poly-0.01 &          0.0155 &          0.0005 &    600.1181 &         0.0127 &         0.0003 &     0.1033 \\
 svm &                resnext101-sigmoid-0.01 &          0.0155 &          0.0005 &    601.0287 &         0.0127 &         0.0003 &     0.1034 \\
 svm &                 resnext101-poly-0.0001 &          0.0155 &          0.0005 &    599.4391 &         0.0127 &         0.0003 &     0.1034 \\
 svm &                    resnext101-rbf-0.01 &          0.0155 &          0.0005 &    602.0599 &         0.0127 &         0.0003 &     0.1037 \\
 svm &                  resnext101-rbf-0.0001 &          0.0155 &          0.0005 &    601.5051 &         0.0127 &         0.0003 &     0.1038 \\
 svm &              resnext101-sigmoid-0.0001 &          0.0155 &          0.0005 &    600.7526 &         0.0127 &         0.0003 &     0.1039 \\
\end{longtable}

% % use section* for acknowledgment
% \section*{Acknowledgment}


\end{document}